\documentclass[journal]{IEEEtran}
\usepackage{multirow}
\usepackage{amssymb}
\usepackage{booktabs}
\usepackage{graphicx}
\usepackage{amsmath}
\usepackage{makecell}
\usepackage{threeparttable}
\usepackage{placeins}
\usepackage{xcolor}

\ifCLASSINFOpdf
\else
\fi

\usepackage{multirow}
\usepackage[numbers,sort&compress]{natbib}

\begin{document}

\title{HiCD: Change Detection in Quality-Varied Images via Hierarchical Correlation Distillation}

\author{Chao~Pang,
        Xingxing~Weng,
        Jiang~Wu,
        Qiang~Wang,
        and~Gui-Song~Xia
\thanks{C. Pang, X. Weng and G.-S. Xia are with School of Computer Science, Wuhan University, Wuhan {\rm 430072}, China. }
\thanks{J. Wu is with Shanghai Artificial Intelligence Laboratory, Shanghai, China. }
\thanks{Q. Wang is with SenseTime Research, Beijing, China. }
\thanks{\textit{Corresponding Author: Gui-Song Xia (guisong.xia@whu.edu.cn)}}
}


\maketitle

\begin{abstract}

Advanced change detection techniques primarily target image pairs of equal and high quality. However, variations in imaging conditions and platforms frequently lead to image pairs with distinct qualities: one image being high-quality, while the other being low-quality. These disparities in image quality present significant challenges for understanding image pairs semantically and extracting change features, ultimately resulting in a notable decline in performance.
To tackle this challenge, we introduce an innovative training strategy grounded in knowledge distillation. The core idea revolves around leveraging task knowledge acquired from high-quality image pairs to guide the model's learning process when dealing with image pairs that exhibit differences in quality. Additionally, we develop a hierarchical correlation distillation approach (involving self-correlation, cross-correlation, and global correlation). This approach compels the student model to replicate the correlations inherent in the teacher model, rather than focusing solely on individual features. This ensures effective knowledge transfer while maintaining the student model's training flexibility.
Through extensive experimentation, we demonstrate the remarkable superiority of our methodologies in scenarios involving only resolution disparities, single-degradation, and multi-degradation quality differences.
The codes will be released at \textit{https://github.com/fitzpchao/HiCD}.

\end{abstract}


\IEEEpeerreviewmaketitle

\section{Introduction}
\IEEEPARstart{C}{hange} detection (CD), which aims to identify changes in Earth's surface using bi-temporal remote sensing image pairs captured within the same area, holds immense significance in numerous applications, including urban planning~\cite{howarth1983landsat,viana2019land,hegazy2015monitoring} and disaster assessment~\cite{zheng2021building,washaya2018coherence,trinder2012aerial,sophiayati2009onboard}, among others. To date, numerous initiatives have been made to develop automated and precise change detection methods. Recently, deep learning-based methods~\cite{daudt2018fully,liu2020building,changeformer} have attracted a lot of interest due to their outstanding performance. 

\begin{figure}[t!]
    \centering
    \includegraphics[width=0.95\linewidth]{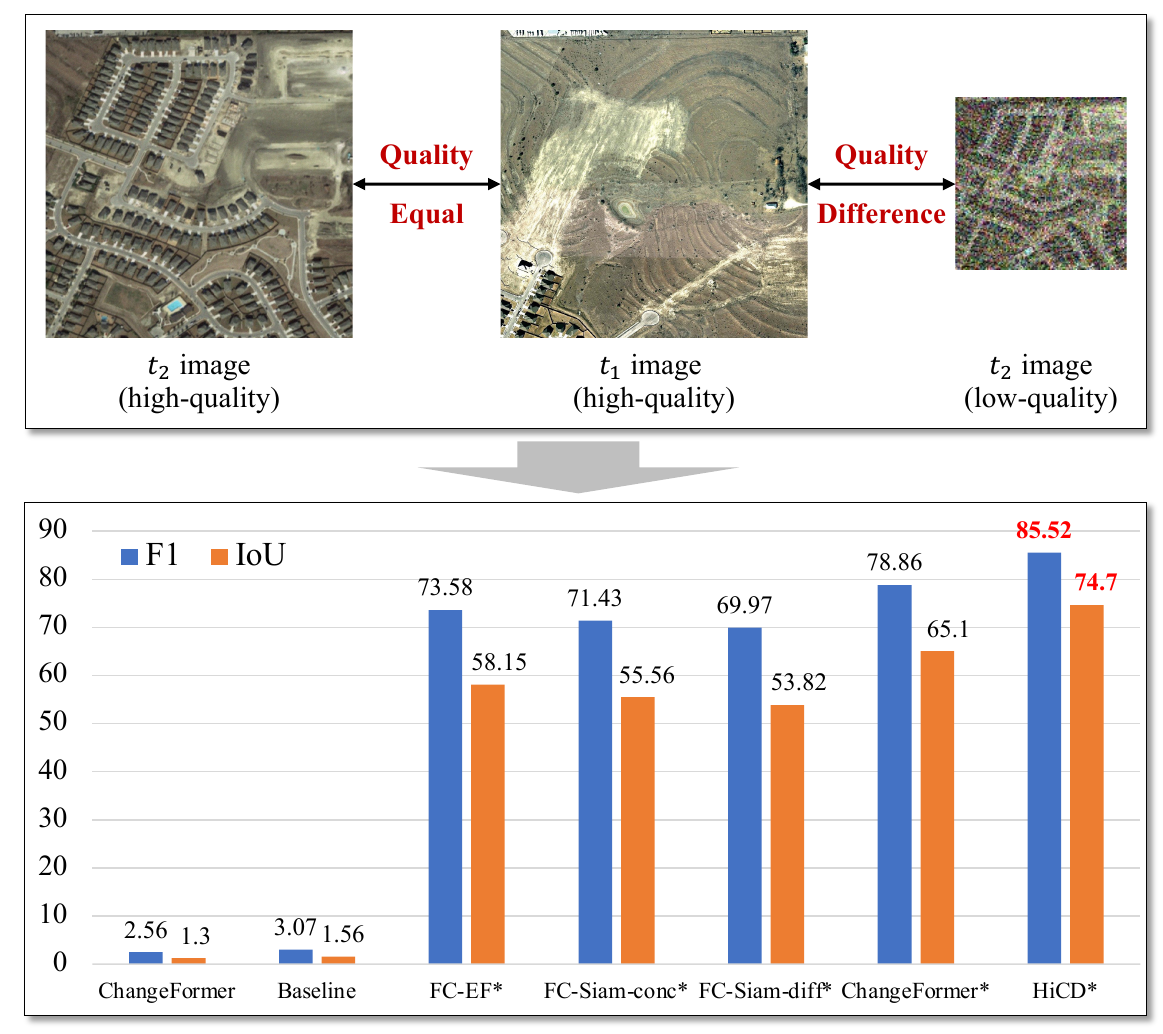}
    \caption{The performance of the excellent CD methods on the multi-degradation LEVIR-CD dataset. * indicates the method trained on quality-varied image pairs, otherwise on the equal high-quality image pairs.}
    \label{fig: motivation}
\end{figure}

However, it is worth noting that the majority of deep learning-based change detection methods~\cite{daudt2018fully,liu2020building,changeformer, BIT} excel in handling pairs of images with equal and high quality, as exemplified by the top-left two images in Figure~\ref{fig: motivation}. 
Nonetheless, they often face significant performance degradation when tasked with scenarios where bi-temporal images exhibit varying qualities, as seen in the top-right image pair in Figure~\ref{fig: motivation}.
The variation in image quality between bi-temporal images primarily stems from fluctuations in imaging conditions, such as the presence of clouds, fog, haze, or smoke, during the acquisition of these bi-temporal image pairs. 
Additionally, random signals introduced during processes such as storage, compression, and transmission can also result in image pairs displaying varying quality, leading to disparities in noise levels between the bi-temporal images. Moreover, in certain real-world applications, such as disaster monitoring, there is a need to utilize image pairs obtained from different platforms with differing spatial resolutions. Consequently, there is a compelling need for the development of change detection (CD) methods that can effectively account for these variations in quality within bi-temporal image pairs.

Up to this point, several studies~\cite{wang2022spcnet, tian2022racdnet, Liu2022g, liu2022learning, chen2023remote} have been dedicated to addressing change detection tasks involving image pairs with differences in spatial resolution. These studies have reported promising results, achieved either by developing methods to directly extract high-resolution features from low-resolution images or by exploiting the advanced methods in super-resolution to enhance low-resolution images before conducting change detection. Nevertheless, there remains a deficiency of effective methods capable of handling change detection tasks involving a broader spectrum of quality differences mentioned earlier.

Different from existing works that only focus on image pairs with different resolutions, quality differences concern diverse degradation factors such as blur, noise, or combinations of them. As a result, the loss and damage of ground object information in low-quality images is more severe than that in low-resolution images. Thus, it is difficult for existing methods to precisely perceive the semantics of low-quality images. Moreover, image pairs with different resolutions provide different delicate levels of details of the same semantic object, which poses challenges to feature alignment of bi-temporal images and consequently affects the final change identification. In change detection with quality-varied image pairs, the problem of feature alignment becomes more difficult because the combination of various degradations leads to a huge gap in both local details and global appearance between objects with the same semantics.


A simple solution to tackle the quality difference challenge is to first restore low-quality images and then perform change detection using restored image pairs~\cite{he2019spectral}. However, the performance of such a solution heavily relies on image restoration methods and can easily be restricted due to error accumulation. Inspired by the fact that CD models trained on high-quality image pairs have captured valuable deep priors that benefit change detection tasks, we propose to leverage the deep priors to guide the representation learning and feature alignment of the CD model on image pairs with different quality.

To this end, we design semantic feature distillation (SFD) and change feature distillation (CFD) modules, based on knowledge distillation. Specifically, the SFD-module concerns how the student model extracts good feature representation from low-quality images, while the CFD-module answers the question of how the student model aligns quality-varied feature pairs and obtains essential change features. During the training, a teacher model for change detection tasks is first trained on high-quality image pairs. Then, we optimize the target model (\emph{i.e.}, the student) on quality-varied pairs composed of high-quality and low-quality images. In parallel, the teacher model takes the corresponding high-quality image pairs as input to extract bi-temporal image representations and change features, which are used to guide the student model via the SFD and CFD modules. 

Traditional feature distillation~\cite{heo2019comprehensive} encourages the student model to mimic representations generated by the teacher model in a manner of individual feature alignment. However, when images suffer from quality degradation, it inevitably results in loss and damage to object information. Strictly aligning individual features between the teacher and student models is extremely challenging and may lead to negative constraints. We argue that the correlation between objects within images is invariant even if the image degrades. Thus, we present a correlational feature distillation that transfers pixel-to-pixel correlations. Notably, we design self-correlation distillation and cross-correlation distillation. The former transfers pixel correlations within an image, improving the understanding capability of the student model for low-quality images. The latter focuses on transferring pixel correlations among bi-temporal images, guiding the student model to highlight discriminative features relevant to change detection tasks. 


Moreover, previous studies~\cite{wang2021exploring,zhou2022rethinking,yang2022cross} have found that global semantic information across the whole training set is of great significance in many computer vision tasks, such as semantic segmentation~\cite{wang2021exploring}. Building upon this insight, we extend cross-correlation to global semantic correlations across the whole training image pair, enabling the student model to better capture key features for change detection tasks. In general, we propose correlational feature distillation to transfer hierarchical pixel correlations learned by the teacher model, \emph{i.e.}, self-correlation, cross-correlation and global correlation, and thus design the SFD-module and the CFD-module to overcome challenges of representation learning and feature alignment caused by quality differences. We denote our method as HiCD, namely \emph{\textbf{Hi}erarchical \textbf{C}orrelation \textbf{D}istillation for quality-varied \textbf{C}hange \textbf{D}etection}.


The proposed HiCD has been evaluated with three experimental settings: only resolution difference, single-degradation quality difference and multi-degradation quality difference, and achieves state-of-the-art results on three popular benchmarks for change detection. Specifically, in the only resolution difference setting, HiCD exceeds existing methods by $\mathbf{3.47\%}$ on the LEVIR-CD dataset~\cite{chen2020spatial}, $\mathbf{5.33\%}$ on the BANDON dataset~\cite{pang2023detecting}, and $\mathbf{3.57\%}$ on the SV-CD dataset~\cite{CDDdataset} in terms of average IoU. For single-degradation quality difference settings, HiCD outperforms all comparative methods with a gap of nearly $\mathbf{3.5\%}$ on LEVIR-CD and BANDON datasets. Multi-degradation quality difference experiment is the most challenging. Most methods reach IoU inferior to $60\%$ on the LEVIR-CD dataset and $30\%$ on the BANDON dataset, while our method is still able to achieve good performance $(\mathbf{74.70\%}$ and $\mathbf{45.74\%})$.


The main contributions of this article are three-fold:
\begin{itemize}
    \item We devise a novel training strategy based on knowledge distillation for addressing the challenges of representation learning and feature alignment caused by quality differences.
    \item We propose a feature distillation for change detection that aims to transfer hierarchical pixel correlations. By forcing the student to mimic the teacher's pixel correlations, we improve the student's capacity to extract good feature representations from low-quality images and mine essential change features from quality-varied representations.
    \item We provide strong performance benchmarks for quality-varied change detection on three widely used datasets: LEVIR-CD, BANDON, and SV-CD.
\end{itemize}

The rest of the article is organized as follows. Section~\ref{Related Work} reviews the related works. Section~\ref{Proposed Method} elaborates on the details of hierarchical correlation distillation for quality-varied change detection. Section~\ref{Experiments} reports experimental results and provides analysis. Finally, conclusions are drawn in Section~\ref{Conclusion}.

\section{Related Work}
\label{Related Work}
\subsection{General Change Detection}
Deep learning has attained remarkable success in change detection for remote sensing images. Numerous advanced deep learning techniques, like deep convolutional neural networks (CNN)~\cite{he2016deep} and fully convolutional networks (FCN)~\cite{long2015fully}, have been extensively employed to devise accurate change detection models. Since change detection usually has two images as inputs, the Siamese network is introduced to process bi-temporal images in parallel \cite{zhan2017change}, then the concatenation or difference operations \cite{daudt2018fully} are used to fuse features of image pairs, and extract change features. Following such Siamese structure, there are two groups of work to boost performance: improving feature representation capability for image pairs and enhancing the interaction between bi-temporal features. 

In detail, advanced backbones such as HRNet~\cite{wang2020deep}, are introduced to extract discriminative features~\cite{Liang2022high,liu2020building}. As the global feature representation is more generalized and discriminative than the local feature representation, dilated convolution~\cite{chen2019spatioc} and attention mechanisms~\cite{Zhang2020,chen2020dasnet} are employed to capture the global context. Feature interaction~\cite{feng2022icif,fang2023changer} is of significant process in change detection tasks, which aims to fuse bi-temporal features and mine change features to produce final change maps. Thus, how to enhance feature interaction has drawn attention in the past few years. Attention operation is a useful tool to achieve feature interaction and has been widely used to replace the simple difference operation~\cite{Zhang2020,chen2020dasnet,liu2020building,chen2020spatial,shi2021deeply}. Recently, an impressive work~\cite{fang2023changer} devises alternative interaction layers in feature extraction to emphasize the effect of feature interactions. Besides, some works~\cite{pang2023detecting, zheng2022changemask} explore feature interaction between semantic segmentation and change detection, and show the potential for performance improvement.

Transformers~\cite{vaswani2017attention} have currently drawn significant attention in the research on change detection due to their superiority in modeling long-range dependencies. BIT~\cite{BIT} and ChangeFormer~\cite{changeformer} are representative examples. Specifically, BIT~\cite{BIT} employs the transformer encoder in the interaction process of bi-temporal features, to efficiently model contexts within the spatial-temporal domain. ChangeFormer~\cite{changeformer} unifies the hierarchical transformer encoder with multi-layer perception decoder to render multi-scale long-range details required for change detection. To fully exploit the powerful global information modeling capabilities of transformers, the change detection model based on pure transformer has been proposed~\cite{swinsunet}. Despite the promising results obtained so far, the above models fail to handle change detection with quality-varied image pairs.

\subsection{Quality-varied Change Detection}
Current research on quality-varied change detection focuses largely on resolution differences. Existing methods can generally be divided into two groups: image restoration-based methods and feature alignment-based methods. The image restoration-based methods~\cite{Liu2022g,tian2022racdnet} aim to reconstruct the natural and detailed high-resolution image from the low-resolution image via off-the-shelf super-resolution models or additional super-resolution modules. For instance, SRGAN~\cite{ledig2017photo} is employed~\cite{Liu2022g} to conduct the super-resolution process on low-resolution images before change detection. Such a way may limit the performance of change detection due to error accumulation. 

The goal of feature alignment-based methods~\cite{chen2023remote,liu2022learning,Shao2021} is to directly align features of bi-temporal images with different resolutions. For example,~\cite{wang2022spcnet} adopts the super-resolution solution to upscale low-resolution feature maps into high-resolution maps. The scale-invariant learning method is proposed~\cite{chen2023remote} to enable the model to adapt to continuous resolution difference ratios. They degrade the high-resolution image by random downsampling to reduce the gap between high-resolution and low-resolution images. Then, a change decoder with implicit neural representation is used to generate high-resolution change maps. 

In addition to the above supervised methods, \cite{zheng2021unsupervised} achieves cross-resolution change detection through an unsupervised method that detects changes by measuring distances between image regions where pixels reside, based on the image segmentation. This method shows potential in addressing the resolution difference issue but struggles to distinguish changes of interest (\emph{e.g.}, building changes) from pseudo changes (caused by seasonal and temporary objects), resulting in a number of false alarms.

Although these methods report impressive results on resolution-difference change detection, they still do not perform well on handling quality differences. Due to the diversity of real imaging conditions and platforms, quality differences of bi-temporal images can be caused by various factors, which lead to challenges for existing methods.


\begin{figure*}[htb!]
    \centering
    \includegraphics[width=0.9\textwidth]{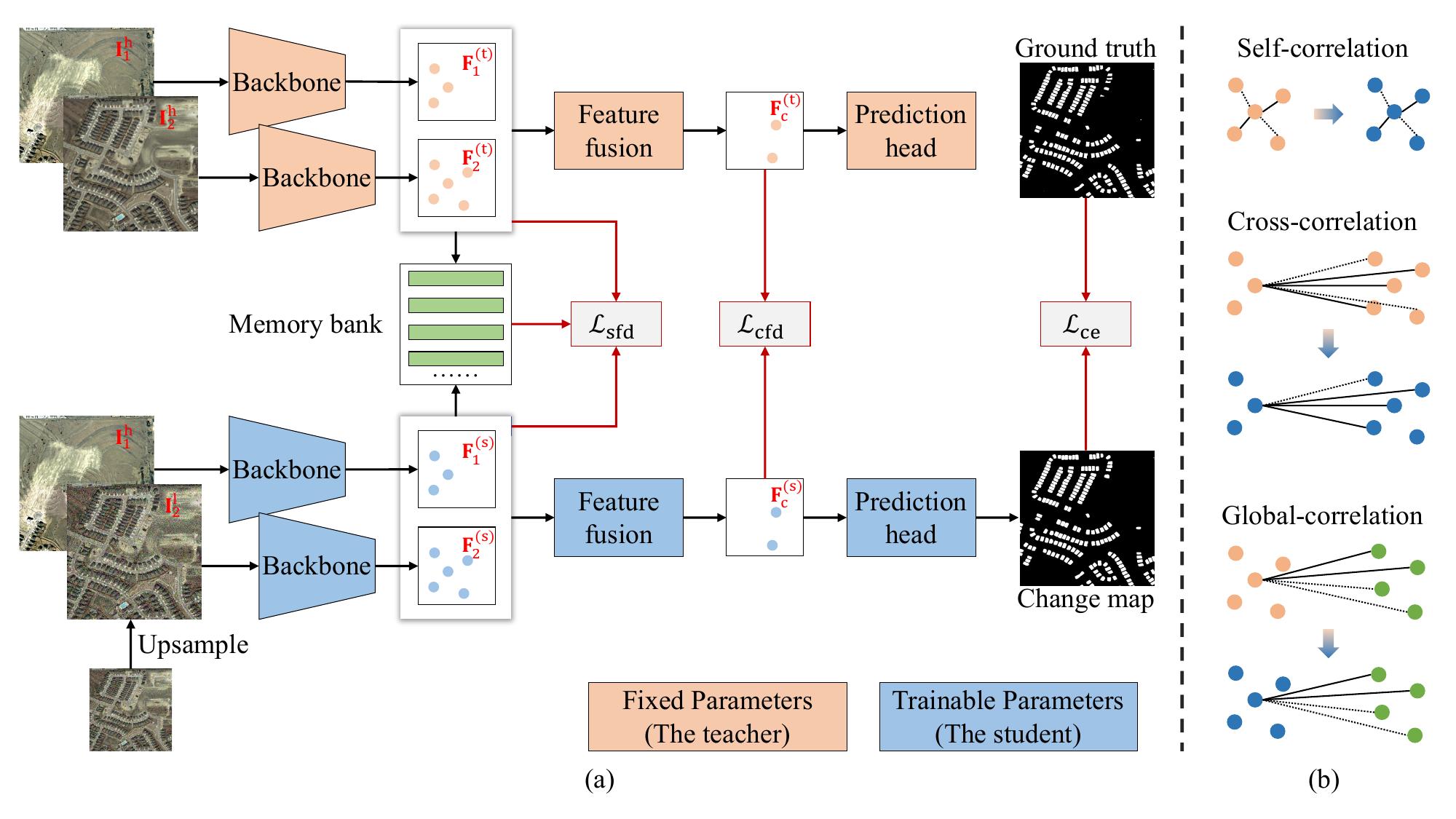}
    \caption{The overview of our proposed method. (a) Knowledge distillation-based training strategy. (b) Hierarchical correlation distillation. During the training of the student, the teacher takes the corresponding high-quality image pairs as input, to generate bi-temporal representations and the change feature for guiding the student. Based on the novel feature distillation for change detection, \emph{i.e.}, hierarchical correlation distillation, we devise the semantic feature distillation module and the change feature distillation module, to transfer different kinds of knowledge acquired by the teacher. $\mathcal{L}_\mathrm{ce}$, $\mathcal{L}_\mathrm{sfd}$ and $\mathcal{L}_\mathrm{cfd}$ are cross-entropy loss, distillation losses of the SFD and CFD modules, respectively.}
    \label{fig: training strategy}
\end{figure*}

\section{Proposed Method}
\label{Proposed Method}
\subsection{Problem Definition}
Given a pair of images $\mathbf{I}_{1}, \mathbf{I}_{2}$ with size of $H \times W$ and a number of $C$ channels, taken from the same geographical area at different times, $t_1$ and $t_2$, the task of change detection aims to produce a change map $\mathbf{Y} \in \left\{0,1 \right\}^{H \times W}$, where each pixel indicates whether the same position of the inputs are changed or not. In this article, $0$ represents \textit{non-change} while $1$ refers to  \textit{change}. The variations in imaging conditions and platforms often lead to quality differences between $\mathbf{I}_{1}$ and $\mathbf{I}_{2}$, which consequently affects the performance of most existing CD methods. Thus, quality-varied change detection aims to generate an accurate change map from bi-temporal inputs with different image quality, \emph{i.e.}, a high-quality image (hq-image) and a low-quality image (lq-image).

Since there are no open-source CD datasets considering quality differences, we adopt a degradation model to synthesize lq-image from hq-image of a certain phase, achieving bi-temporal quality differences. In this article, we uniformly perform image degradation on hq-images acquired at time $t_2$. For clarity, we modify the superscript of bi-temporal image pairs, \emph{i.e.}, $\mathbf{I}_{1}^\mathrm{h}$ and $\mathbf{I}_{2}^\mathrm{l}$. Different from current research only focuses on the resolution difference, we take into account diverse degradation types. The degradation model is mathematically modeled by:
\begin{equation}
\label{degradation-model}
 \mathbf{I}_{2}^\mathrm{l}=(\mathbf{I}_{2}^\mathrm{h} \otimes \mathbf{k})
 \downarrow_{\mathbf{s}}+\mathbf{n},
\end{equation}    
where $\mathbf{k}$ and $\otimes$ denote the Gaussian kernel and convolution operation, respectively, used for image blurring. $\downarrow_{\mathbf{s}}$ represents a downsampling operation with a scale factor of $\mathbf{s}$. $\mathbf{n}$ accounts for white Gaussian noise. This degradation model assumes that the lq-image is obtained by subjecting the hq-image to a sequence of operations including blur, noise addition, and downsampling. It is essential to acknowledge that the real-world degradation of remote sensing images encompasses various intricate factors. While our method is trained on commonly observed degradations, we argue that the proposed change detection model can effectively address quality differences arising from unseen degradation scenarios, as indicated by the experimental results in Table~\ref{tab:levir_single_deg}~(a) and Table~\ref{tab:bandon_single_deg}~(b).

\subsection{Knowledge Distillation-Based Training Strategy}

To handle the quality difference challenge, we propose a knowledge distillation-based training strategy for change detection, namely HiCD. The key idea is to leverage deep priors learned from hq-image pairs to guide representation learning and feature alignment of a CD model on high-quality and low-quality (\emph{i.e.}, hq-lq) image pairs. Figure \ref{fig: training strategy} illustrates the overall training process. There is a teacher model and a student model with the same architecture. Following the mainstream change detection framework, the teacher (or student) consists of three core components, a weight-sharing Siamese backbone for representation learning of bi-temporal images, a feature fusion module that aligns bi-temporal representations to mine the change feature, and a prediction head to achieve the change map. 

During the training, the teacher is first trained on hq-image pairs and frozen. Then, the student takes hq-lq image pairs as input for model optimization. Notably, to make the student adapt to the lq-image regardless of resolution, we upsample the lq-image $\mathbf{I}_{2}^\mathrm{l}$ to the size of the hq-image $\mathbf{I}_{1}^\mathrm{h}$, via bicubic interpolation. Meanwhile, the teacher takes the corresponding hq-hq image pairs as input to generate bi-temporal representations and the change feature, which are used to guide the student. To implement this, we present a semantic feature distillation (SFD) module and a change feature distillation (CFD) module for transferring different kinds of knowledge from the teacher. Specifically, the SFD-module utilizes the teacher's feature representations to guide the student on extracting similar high-quality feature representations from lq-image, while the CFD-module uses the teacher's change features as the distilling goal, teaching the student how to align high-quality and low-quality representations, and then obtain essential change features for final change map prediction. 

The overall objective $\mathcal{L}$ of the student model is defined as:
\begin{equation}
\mathcal{L} = \mathcal{L}_\mathrm{ce}+\lambda_\mathrm{sfd} \cdot \mathcal{L}_\mathrm{sfd} + \lambda_\mathrm{cfd} \cdot \mathcal{L}_\mathrm{cfd},
\end{equation}    
where $\mathcal{L}_\mathrm{ce}$ is cross-entropy loss for change detection tasks, $\mathcal{L}_\mathrm{sfd}$ and $\mathcal{L}_\mathrm{cfd}$ are distillation losses of semantic features and change features, respectively. $\lambda_\mathrm{sfd}$ and $\lambda_\mathrm{cfd}$ are tunable hyperparameters to balance the loss terms. In subsequent sections, we first explain the novel feature distillation approach for change detection and then show the definition of $\mathcal{L}_\mathrm{sfd}$ and $\mathcal{L}_\mathrm{cfd}$.

\subsection{Hierarchical Correlation Distillation}
A common way of feature distillation is directly minimizing individual feature differences between the teacher and the student in a pixel-wise manner. However, due to image quality degradation and ill-posed problems of feature restoration, it is difficult to align individual features, especially in the edge regions and even causes negative constraints. We argue that correlations between geospatial objects are more stable than semantic features of individual objects when image quality keeps degrading. On this basis, we propose to transfer correlations inside features learned by the teacher rather than individual features themselves. Correlation distillation does not force the student to match the teacher's feature maps directly but encourages the student to focus on correlations inside features, giving more flexibility to the student in training. For its concrete realizations, we present hierarchical correlation distillation: self-correlation, cross-correlation, and global-correlation distillations. Specifically,

\textbf{Self-correlation Distillation.} Exactly perceiving semantic information inside lq-image is challenging due to the damage and loss of object information caused by blur, noise, and downsampling. The goal of self-correlation distillation is to supervise the student by object correlations within an hq-image, hoping that the understanding capability of the student for lq-image can be improved. Given a teacher model $t$ and a student model $s$, we let $(\mathbf{F}_\mathrm{1}^\mathrm{(t)}, \mathbf{F}_\mathrm{2}^\mathrm{(t)})$ and $(\mathbf{F}_\mathrm{1}^\mathrm{(s)}, \mathbf{F}_\mathrm{2}^\mathrm{(s)})$ be semantic feature representations of bi-temporal inputs of the teacher and the student, respectively, where all feature sizes are $h \times w \times d$. Note that $\mathbf{F}_\mathrm{2}^\mathrm{(t)}$ is obtained from $\mathbf{I}_\mathrm{2}^\mathrm{h}$ by the teacher's backbone, and $\mathbf{F}_\mathrm{2}^\mathrm{(s)}$ is generated from the upsampled $\mathbf{I}_\mathrm{2}^\mathrm{l}$ by the student's backbone. Following~\cite{yang2022cross}, we exploit the cosine similarity between pixels to constitute object correlations within individual images. 

In our implementation, for a feature map such as $\mathbf{F}$ with a size of $h \times w \times d$, we first reshape $\mathbf{F}$ to the size of $hw \times d$. Given a matrix $A \in \mathbb{R}^{h \times w \times d}$, we can use $\mathcal{R}_{hw \times d}(A)$ to denote the function that reshapes matrix $A$ into a new one with size of ${hw \times d}$. Subsequently, we perform channel-wise ${\ell}_{2}$ normalization.  With a matrix $A \in \mathbb{R}^{hw \times d}$, we can represent the normalization function along the channel dimension as $\mathcal{N}$, and this process is described by:
\begin{equation}
    \mathcal{N}(A) = A \oslash (A \otimes A \times J)^{\circ \frac{1}{2}}
\end{equation}
where $J$ represents the all-ones matrix ($J\in\mathbb{R}^{d \times 1}$), $\oslash$ signifies element-wise division, $\otimes$ signifies element-wise multiplication, and $A^{\circ n}$ represents raising each element of matrix $A$ to the power of $n$. Finally, we calculate the matrix multiplication of the normalized feature map and its transpose to compute the self-correlation $\textrm{CorS}(\mathbf{F})$ within $\mathbf{F}$:
\begin{align}
\label{self-c}
\textrm{CorS}(\mathbf{F}) =  \mathcal{N}(\mathcal{R}_{hw \times d}(\mathbf{F})) \times {\mathcal{N}(\mathcal{R}_{hw \times d}(\mathbf{F}))}^{T},
\end{align}    
where $\times$ denotes the matrix multiplication operator, and $A^T$ indicates the transpose of matrix $A$. The dimensions of $\textrm{CorS}(\mathbf{F})$ are $ hw \times hw $.

Utilizing the self-correlations derived from both the teacher and the student, a self-correlation distillation loss $\mathcal{L}_{\mathrm{s2}}$ is formulated as follows:
\begin{align}
\label{self-loss}
\mathcal{L}_{\mathrm{s2}}= \frac{1}{N^2} \sum_{i=1}^N \sum_{j=1}^N\| {\textrm{CorS}(\mathbf{F}_2^{(s)})}_{ij} - {\textrm{CorS}(\mathbf{F}_2^{(t)})}_{ij} \|_2^2,
\end{align}
where $N$ denotes the number of pixels in the rows or columns of the correlation matrix, \emph{i.e.}, $N=hw$, $\|\cdot\|_2$ represents the ${\ell}_{2}$-norm, $(ij)$ signifies the position of the $i$-th row and $j$-th column in the correlation matrix, with $i,j\in\{1,2, \cdots, N\}$. $\mathcal{L}_\mathrm{s2}$ transfers the object correlation of the hq-image to help the student perceive the corresponding lq-image.

Since the student adopts a weight-sharing Siamese backbone to extract feature representations of hq-lq image pair, $\mathcal{L}_\mathrm{s2}$ may make the student customized for the lq-image $\mathbf{I}_\mathrm{2}^\mathrm{l}$ and ignore the hq-image $\mathbf{I}_\mathrm{1}^\mathrm{h}$. Thus, we also transfer the self-correlation inside $\mathbf{F}_\mathrm{1}^{(t)}$. For easy distinguishing, we denote self-correlation distillation loss $\mathcal{L}_\mathrm{s1}$, $\mathcal{L}_\mathrm{s2}$ for representation learning of the student's hq-input and lq-input, respectively. The  $\mathcal{L}_\mathrm{s1}$ can be computed by: 
\begin{align}
\mathcal{L}_{\mathrm{s1}}= \frac{1}{N^2} \sum_{i=1}^N \sum_{j=1}^N\| {\textrm{CorS}(\mathbf{F}_1^{(s)})}_{ij} - {\textrm{CorS}(\mathbf{F}_1^{(t)})}_{ij} \|_2^2.
\end{align}

\begin{figure}[htb!]
    \centering
    \includegraphics[width=\linewidth]{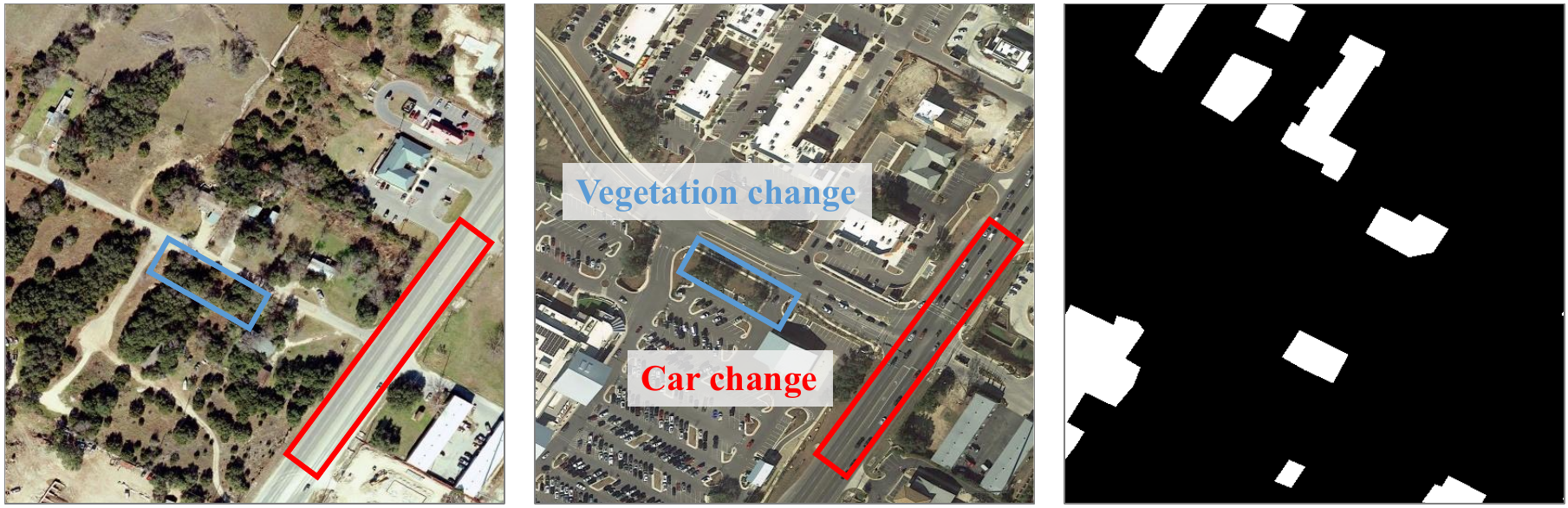}
    \caption{The building change detection is affected by various irrelevant changes.}
    \label{fig: irrelevant change}
\end{figure}

\textbf{Cross-correlation Distillation.} Remote sensing images often with large geographic coverage and there are abundant geospatial objects inside, may lead to a negative impact on change detection tasks. For instance, it can be observed from Figure \ref{fig: irrelevant change} that there are changes caused by seasonal and temporary objects (e.g., vegetation and cars). These changes are not significant in most application scenarios, such as the building change detection shown in Figure~\ref{fig: irrelevant change}, and they can interfere with the detection of meaningful changes of interest. Hence, the model must focus on the key objects strongly related to change detection, instead of precisely depicting the whole image content. Considering this point, we propose to use correlations among bi-temporal image pairs to guide the student in highlighting task-specific representations and contribute to subsequent bi-temporal alignment, which is called cross-correlation distillation.

Similar to self-correlation distillation, the cosine similarity between pixels from different images is used to represent cross-correlation. Then, we utilize the difference in cross-correlations of the teacher and the student to penalize the representation learning of the student. Given two feature maps,  $\mathbf{F}_{1}$, $\mathbf{F}_{2} \in \mathbb{R}^{h \times w \times d}$, the cross-correlations $\textrm{CorC}(\mathbf{F}_{1},\mathbf{F}_{2})$ are computed as follows:
\begin{align}
\resizebox{0.9\linewidth}{!}{$\displaystyle
\textrm{CorC}(\mathbf{F}_{1},\mathbf{F}_{2}) = \mathcal{N}(\mathcal{R}_{hw \times d}(\mathbf{F}_{1})) \times {\mathcal{N}(\mathcal{R}_{hw \times d}(\mathbf{F}_{2}))}^{T},
$}
\label{equ:croc}
\end{align}
where $\textrm{CorC}(\mathbf{F}_{1},\mathbf{F}_{2}) \in \mathbb{R}^{hw \times hw}$. with Equation~\eqref{equ:croc}, we derive the cross-correlations for both the teacher and the student, denoted as  $\textrm{CorC}(\mathbf{F}_{1}^{(t)},\mathbf{F}_{2}^{(t)}) $ and $\textrm{CorC}(\mathbf{F}_{1}^{(s)},\mathbf{F}_{2}^{(s)}) $. Consequently, we formulate a cross-correlation distillation loss as follows:
\begin{equation}
\small
\mathcal{L}_\mathrm{c}= 
\frac{1}{N^2} \sum_{i=1}^N \sum_{j=1}^N\| {\textrm{CorC}(\mathbf{F}_1^{(s)},\mathbf{F}_2^{(s)})}_{ij} - {\textrm{CorC}(\mathbf{F}_1^{(t)},\mathbf{F}_2^{(t)})}_{ij} \|_2^2.   
\end{equation}

\textbf{Global-correlation Distillation.} Except for the self-correlation and cross-correlation, we argue other correlations can also boost the student. For example, one of bi-temporal inputs for the teacher and the student is the same image, \emph{i.e.}, $\mathbf{I}_\mathrm{1}^\mathrm{h}$. Note again that another input for the teacher is $\mathbf{I}_\mathrm{2}^\mathrm{h}$ and for the student is $\mathbf{I}_\mathrm{2}^\mathrm{l}$. If deep priors about change detection have been transferred to the student, the correlation among $\mathbf{F}_\mathrm{1}^\mathrm{(s)}$ and $\mathbf{F}_\mathrm{2}^\mathrm{(t)}$, or $\mathbf{F}_\mathrm{1}^\mathrm{(t)}$ and $\mathbf{F}_\mathrm{2}^\mathrm{(s)}$ should be very similar. Besides, recent research~\cite{wang2021exploring,zhou2022rethinking,yang2022cross} on computer vision tasks demonstrates that the global semantic information across the whole training dataset is valuable and can boost performance. The above points motivate us to extend cross-correlation among registered image pairs to the whole training dataset, \emph{i.e.}, global-correlation. We aim to further enhance the student by penalizing global-correlation differences between it and the teacher.

At each iteration of model training, it is computationally expensive to enumerate all potential correlations to capture global correlation. Inspired by~\cite{yang2022cross}, we adopt a memory bank to store pixel features sampled from $\mathbf{F}_\mathrm{1}^\mathrm{(t)}$, $\mathbf{F}_\mathrm{2}^\mathrm{(t)}$ and $\mathbf{F}_\mathrm{1}^\mathrm{(s)}$ at each iteration. We argue that the quality of $\mathbf{F}_\mathrm{2}^\mathrm{(s)}$ is poor, and it is not suitable for training as a guiding feature. Then, the global correlations of the teacher and the student are obtained by calculating pixel similarity between stored features and outputs of their Siamese backbone. To be specific, let $\mathcal{Q} \in \mathbb{R}^{N_q \times d}$ be the memory bank, where $N_q$ is number of the stored pixel. For each image pair in the batch, we sample a small amount, \emph{i.e.}, $N_p (N_p \ll N_q)$, of pixel features from randomly selected a feature map from $\mathbf{F}_\mathrm{1}^\mathrm{(t)}$, $\mathbf{F}_\mathrm{2}^\mathrm{(t)}$ and $\mathbf{F}_\mathrm{1}^\mathrm{(s)}$, and push them into $\mathcal{Q}$. Like~\cite{he2020momentum}, the memory bank is progressively updated in a first-in-first-out manner. 

During the distillation, $K_q$ pixel features are sampled from $\mathcal{Q}$ and concatenated along the row dimension, resulting in $\mathbf{F}_q \in \mathbb{R}^{K_q \times d}$. Given a feature map $\mathbf{F} \in \mathbb{R}^{ h\times w \times d} $, the global-correlations $\textrm{CorG}(\mathbf{F})$ are computed as follows:
\begin{align}
\label{equ:corg}
\textrm{CorG}(\mathbf{F}) = \mathcal{N}(\mathcal{R}_{hw \times d}(\mathbf{F})) \times {\mathcal{N}(\mathbf{F}_{q})}^{T}
\end{align}
where $\textrm{CorG}(\mathbf{F}) \in \mathbb{R}^{ hw \times hw}$. Using Equation~\eqref{equ:corg}, we can derive four global-correlations: $\textrm{CorG}(\mathbf{F}_{1}^{(t)})$, $\textrm{CorG}(\mathbf{F}_{2}^{(t)})$, $\textrm{CorG}(\mathbf{F}_{1}^{(s)})$ and $\textrm{CorG}(\mathbf{F}_{2}^{(s)})$. Subsequently, a global-correlation distillation loss $\mathcal{L}_\mathrm{g}$ is defined as follows:
\begin{equation}
\begin{aligned}
 \mathcal{L}_\mathrm{g} = \frac{1}{N^2} \sum_{i=1}^N \sum_{j=1}^N\| {\textrm{CorG}(\mathbf{F}_1^{(s)})}_{ij} - {\textrm{CorG}(\mathbf{F}_1^{(t)})}_{ij} \|_2^2 \\ 
 +  \frac{1}{N^2} \sum_{i=1}^N \sum_{j=1}^N\| {\textrm{CorG}(\mathbf{F}_2^{(s)})}_{ij} - {\textrm{CorG}(\mathbf{F}_2^{(t)})}_{ij} \|_2^2.
\end{aligned}  
\end{equation}

\subsection{Feature Distillation Module with Hierarchical Correlation}
Hierarchical correlation distillation composed of self-correlation, cross-correlation, and global correlation, transfers different knowledge acquired by the teacher to the student. These distillation losses can be used either alone or together. In this article, we use all kinds of correlations in the SFD-module due to the complementary between them. Thus, the distillation loss of this module is defined by:
\begin{equation}
\begin{aligned}
\mathcal{L}_\mathrm{sfd}&=\lambda_\mathrm{s1}\mathcal{L}_\mathrm{s1}+\lambda_\mathrm{s2}\mathcal{L}_\mathrm{s2}+\lambda_\mathrm{c}\mathcal{L}_\mathrm{c}+\lambda_\mathrm{g}\mathcal{L}_\mathrm{g},    
\end{aligned}
\end{equation}
where $\lambda_\mathrm{s1}$, $\lambda_\mathrm{s2}$, $\lambda_\mathrm{c}$ and $\lambda_\mathrm{g}$ are tunable hyperparameters to balance the corresponding loss terms.

After generating the bi-temporal feature representations, the teacher (or the student) first concatenates them along the channel dimension and then feeds them into several convolutional layers for feature fusion and extracting the change features $(\mathbf{F}_\mathrm{c}^\mathrm{(t)}$, $\mathbf{F}_\mathrm{c}^\mathrm{(s)}$). As the CFD-module aims to guide the student on how to obtain essential change features, we simply use self-correlation distillation loss in this module. The distillation loss $\mathcal{L}_\mathrm{cdf}$ can be expressed by:
\begin{equation}
\mathcal{L}_\mathrm{cdf} = \frac{1}{N^2} \sum_{i=1}^N \sum_{j=1}^N\| {\textrm{CorS}(\mathbf{F}_{c}^{(s)})}_{ij} - {\textrm{CorS}(\mathbf{F}_{c}^{(t)})}_{ij} \|_2^2 \\ ,
\end{equation}  
where $\textrm{CorS}(\mathbf{F}_{c}^{(t)})$ and $\textrm{CorS}(\mathbf{F}_{c}^{(s)})$ are correlations within the change feature maps of the teacher and the student, respectively.

\section{Experiments}
\label{Experiments}
\subsection{Experimental Setup}\label{section:setting}
\textbf{Datasets.} We compare the proposed method to the state-of-the-art methods on three commonly-used CD datasets: LEVIR-CD~\cite{chen2020spatial}, BANDON~\cite{pang2023detecting} and SV-CD~\cite{CDDdataset}. To be specific, \textit{LEVIR-CD} is a dataset for building change detection, which consists of 637 pairs of high-resolution remote sensing images with size $1024 \times 1024$. As done by~\cite{chen2020spatial}, we split the dataset into 445 pairs for training, 64 pairs for validation, and 128 pairs for testing. \textit{BANDON} is a building CD dataset composed of off-nadir aerial images, which includes 2283 image pairs with size $2048 \times 2048$. Similar to \cite{pang2023detecting}, we use 1689 pairs for training, 202 pairs for validation, and 207 pairs for in-domain testing. \textit{SV-CD} is a large-scale CD dataset that contains 16000 image pairs with size $256 \times 256$. Different from LEVIR-CD and BANDON, this dataset focuses on changes in buildings, cars, and roads. We use the data divided by the official partitioning~\cite{CDDdataset}, with 10,000/3,000/3,000 pairs respectively employed for training/validation/testing.

\begin{figure*}[htb!]
    \centering
    \includegraphics[width=\linewidth]{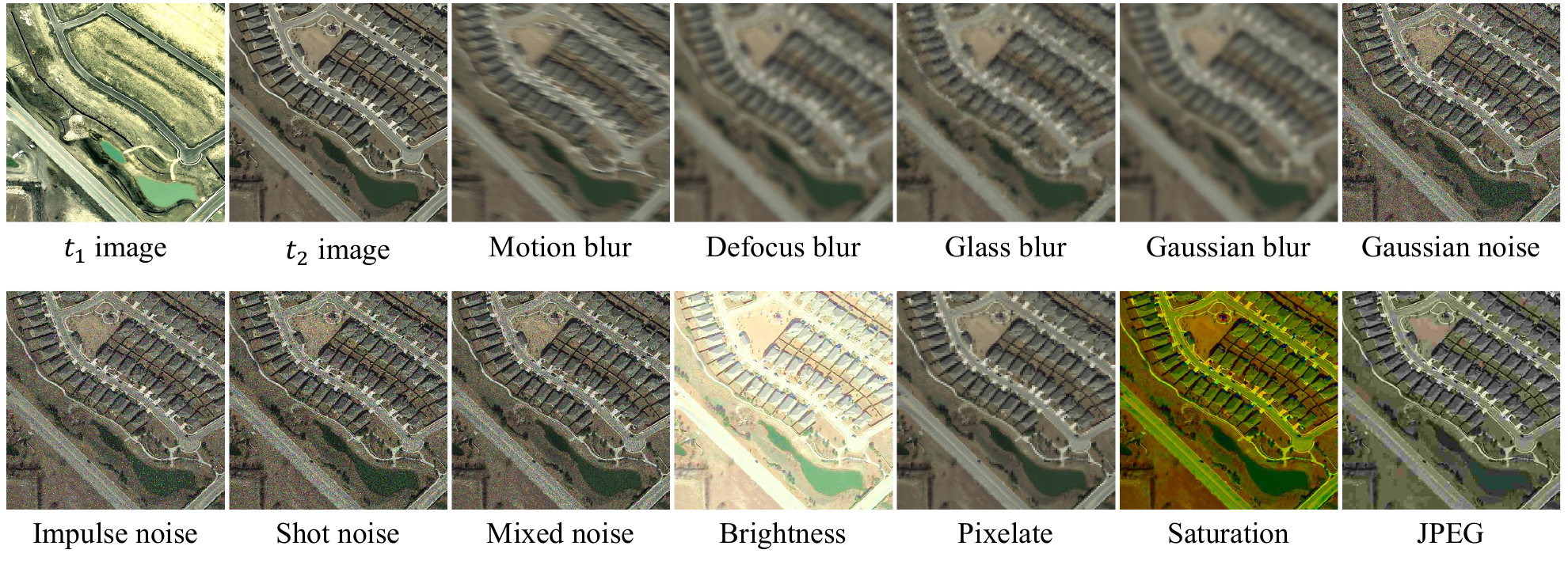}
    \caption{Example of the LEVIR-CD dataset with 12 types of degradations.}
    \label{fig: single_deg_example}
\end{figure*}

\textbf{Experimental Settings.} To fully verify the superiority of our method, we define three experimental settings, depending on how we degrade testing images to build the quality-varied image pairs. Since current research on quality-varied CD only focuses on handling the resolution difference, we define an experimental setting called \textit{only resolution difference}: only the downsample operation (\emph{i.e.}, bilinear interpolation) is used to process the images captured at time $t_2$. Following \cite{chen2023remote}, we gradually downsample testing sets of LEVIR-CD and BANDON by $\left\{1\times,1.3\times,2\times,3\times,4\times,5\times,6\times,8\times\right\}$, SV-CD by $\left\{1\times,2\times,4\times,5\times,8\times,9\times,10\times,12\times\right\}$, to evaluate models' ability to overcome continuous resolution-difference ratios. 

The second setting that we denote as \textit{single-degradation quality difference}: 12 kinds of degradation in~\cite{kamann2020benchmarking} are independently employed to reduce the quality of $t_2$ images. As shown in Figure \ref{fig: single_deg_example}, there are degradation types of motion blur, defocus blur, glass blur, Gaussian blur, Gaussian noise, impulse noise, shot noise, mixed noise, brightness, pixelate, saturation, and JPEG compression. These degradations are only used to process the testing set and each type of degradation has five levels of severity. Please note that, consistent with \cite{xie2021segformer}, only the severity levels of the first three low levels are considered as evaluation metrics for the 4 types of degradation in noise.

We defined the third setting as \textit{multi-degradation quality difference} where $t_2$ images are degraded by blur, noise, and downsampling, simultaneously. The degradation model in Equation \eqref{degradation-model} is used to achieve the multi-degradation quality difference. Following \cite{cui2022exploring}, the downsampling operation with the default scale factor $s=8$  is randomly chosen from bilinear interpolation, nearest neighbor interpolation and bicubic interpolation. The blur kernel has two kinds of isotropic Gaussian and anisotropic Gaussian kernels, with kernel size uniformly sampled from $\{7\times7, 9\times9, ..., 21\times21\}$. The width of the isotropic Gaussian kernel is uniformly chosen from $(0.1, 2.4)$. For the anisotropic Gaussian kernel, the kernel angle can range from $0$ to $\pi$ and the longer kernel width is uniformly chosen from $(0.5, 6)$. Zero-mean Gaussian white noise with variance randomly chosen from a uniform distribution $(0, 25/255)$, is used to simulate sensor noise.

\begin{figure*}[htb!]
    \centering
    \includegraphics[width=\linewidth]{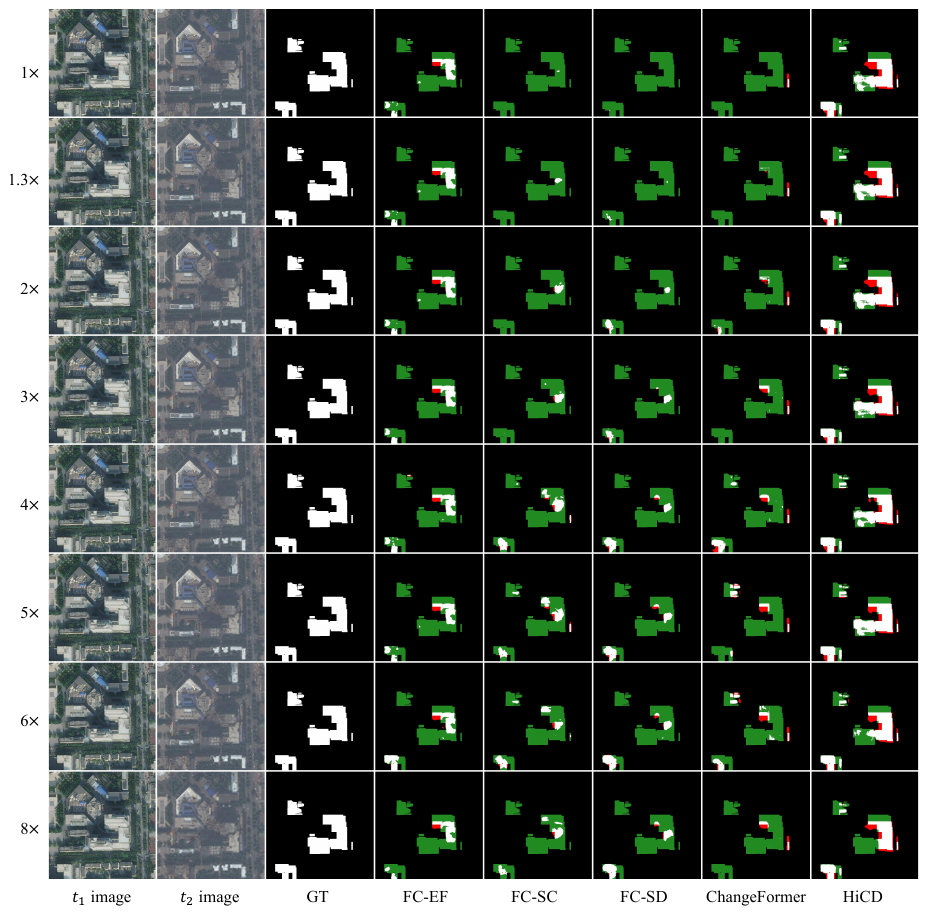}
    \caption{Qualitative results on the BANDON dataset with the setting of only resolution difference. \textit{GT}, \textit{FC-SC}, and \textit{FC-SD} are short for the ground truth, FC-Siam-conc \cite{daudt2018fully} and FC-Siam-diff \cite{daudt2018fully}, respectively. Different colors, \emph{i.e.}, white, black, red, and green, indicate true positive, true negative, false positive, and false negative.}
    \label{fig: resolution_resluts}
\end{figure*}

\textbf{Implementation Details.} We use SegFormer-B2~\cite{xie2021segformer} as the weight-sharing backbone of the teacher and the student, to generate semantic features with size $[\frac{H}{4} \times \frac{W}{4} \times 512]$. Then, bi-temporal features are concatenated and fed into two stacked convolution layers with a BN layer and two ReLU layers, to extract the change feature with size $[\frac{H}{4} \times \frac{W}{4} \times 256]$. These two convolution layers respectively have 256 kernels with the size of $3\times3$. The prediction head comprises two deconvolution layers with ResNet blocks and a convolution layer to yield the final change map. In the deconvolution layer, the kernel size is $4\times4$ with a stride of 2. For training the student, we empirically set $N_q=20480$, $K_q=4096$, $N_p=8$ to calculate the global-correlation distillation, and $\lambda_\mathrm{s1}=1$, $\lambda_\mathrm{s2}=1$, $\lambda_\mathrm{c}=1$, $\lambda_\mathrm{g}=0.4$, $\lambda_\mathrm{sfd}=5$ and $\lambda_\mathrm{cfd}=0.25$ to balance the corresponding loss terms. The images used in this article have three channels, hence $C=3$.

\textbf{Training Details.} The teacher model is trained on training sets augmented by random flip and random rotation, while the student is trained on training sets augmented by the degradation model in Equation \eqref{degradation-model}. The hyperparameters of the degradation model are the same as the above multi-degradation quality difference. For all models, we adopt the AdamW as our optimizer with an initial learning rate of 0.001, $\beta_1=0.9$, $\beta_2=0.999$, and weight decay of 0.01. 
We trained for 40K iterations on LEVIR-CD and SV-CD, and 80K iterations on BANDON. All models were trained on 4 NVIDIA Tesla V100 GPUs and decayed the learning rate using the poly policy with a power of 0.9. The batch size is set to 16 and images are cropped into $256\times256$ before input. For fairness, we retrain some comparative models using public codes with default parameter setup.

\textbf{Comparative Methods.} Since there is rare work on quality-varied change detection, we compare our method with excellent methods of general change detection and resolution-difference change detection. For general change detection, we choose CNN-based methods of FC-EF~\cite{daudt2018fully}, FC-Siam-diff~\cite{daudt2018fully}, FC-Siam-conc~\cite{daudt2018fully}, STANet~\cite{chen2020spatial}, IFNet~\cite{Zhang2020}, SNUNet~\cite{Fang2021}, and Transformer-based methods of BIT~\cite{BIT}, ICIFNet~\cite{feng2022icif}, DMINet~\cite{Feng2023}, ChangeFormer~\cite{changeformer}. For resolution-difference change detection, we choose SRCDNet~\cite{Liu2022g}, SUNet~\cite{Shao2021}, MM-Trans~\cite{liu2022learning} and SILI~\cite{chen2023remote}. They employ different strategies to handle the resolution-difference challenge: SRCDNet uses a GAN-based super-resolution module to restore high-resolution images from low-resolution images, while SUNet, MM-Trans and SILI focus on feature alignment. The baseline of our method has the same architecture as the student model, but is trained with only cross-entropy loss. We use F1 score and intersection over union (IoU) to quantitatively evaluate CD results. 

\subsection{Comparative Experiments}

\begin{table*}[ht!]
\renewcommand{\arraystretch}{1.25}
\centering
\caption{Results (F1/IoU in \%) on the LEVIR-CD dataset with the setting of only resolution difference.}
\begin{threeparttable}
\resizebox{\textwidth}{!}{
\begin{tabular}{c|c|cccccccc|c}
\hline
\multirow{2}{*}{Method} &\multirow{2}{*}{Data AUG.} & \multicolumn{8}{c|}{Different resolution difference ratio} & \multirow{2}{*}{Avg.}  \\ \cline{3-10}
& &$1\times$ & $1.3\times$ & $2\times$ & $3\times$ & $4\times$ & $5\times$ & $6\times$ & $8\times$ \\ \hline
FC-EF \cite{daudt2018fully} &\multirow{13}{*}{$\downarrow+4\times$} & 73.67 / 58.31 & 74.49 / 59.35 & 75.11 / 60.14 & 75.49 / 60.62 & 75.31 / 60.40 & 74.41 / 59.24 & 72.52 / 56.88 & 66.83 / 50.18 & 73.47 / 58.14 \\
FC-Siam-conc \cite{daudt2018fully} && 78.63 / 64.78 & 79.33 / 65.74 & 79.32 / 65.73 & 78.34 / 64.39 & 76.40 / 61.81 & 73.55 / 58.16 & 69.91 / 53.73 & 61.57 / 44.48 & 74.63 / 59.85 \\  
FC-Siam-diff \cite{daudt2018fully} && 75.39 / 60.51 & 76.29 / 61.66 & 76.06 / 61.36 & 74.17 / 58.94 & 70.83 / 54.83 & 66.12 / 49.39 & 60.35 / 43.22 & 46.57 / 30.35 & 68.22 / 52.53 \\
STANet \cite{chen2020spatial} && 42.03 / 26.61 & 47.96 / 31.54 & 55.75 / 38.64 & 57.77 / 40.62 & 50.93 / 34.16 & 36.49 / 22.32 & 17.22 / 9.42 & 3.97 / 2.02 & 39.02 / 25.67\\ 
IFNet \cite{Zhang2020} && 74.24 / 59.03 & 77.77 / 63.62 & 81.02 / 68.10 & 83.48 / 71.65 & 83.73 / 72.01 & 82.10 / 69.64 & 78.39 / 64.47 & 66.24 / 49.52 & 78.37 / 64.75  \\
SNUNet \cite{Fang2021} && 85.13 / 74.11 & 86.77 / 76.62 & 87.52 / 77.81 & 87.24 / 77.36 & 85.11 / 74.09 & 76.63 / 62.12 & 59.56 / 42.41 & 17.66 / 9.68 & 73.20 / 61.77  \\  
BIT \cite{BIT} && 86.28 / 75.86 & 86.42 / 76.09 & 86.66 / 76.47 & 86.67 / 76.48 & 85.68 / 74.94 & 81.06 / 68.16 & 70.40 / 54.33 & 28.73 / 16.78 & 76.49 / 64.89 \\
ICIFNet \cite{feng2022icif} && 86.24 / 75.80 & 86.44 / 76.11 & 86.78 / 76.65 & 86.84 / 76.75 & 86.20 / 75.75 & 83.63 / 71.87 & 78.95 / 65.22 & 59.26 / 42.10& 81.79 / 70.03 \\   
DMINet \cite{Feng2023} && 86.28 / 75.87 & 86.49 / 76.20 & 86.85 / 76.75 & 86.96 / 76.93 & 86.89 / 76.82 & 83.78 / 72.08 & 79.10 / 65.43 & 57.40 / 40.26 & 81.72 / 70.04  \\
SUNet \cite{Shao2021} && 75.51 / 60.65 & 75.53 / 60.69 & 75.67 / 60.86 & 75.98 / 61.27 & 76.08 / 61.40 & 75.70 / 60.90 & 75.01 / 60.02 & 69.96 / 53.80 & 74.93 / 59.95  \\  
SRCDNet \cite{Liu2022g} && 75.87 / 61.12 & 76.46 / 61.89 & 76.77 / 62.30 & 76.30 / 61.68 & 74.17 / 58.94 & 69.38 / 53.12 & 60.13 / 42.99 & 29.23 / 17.12 & 67.29 / 52.40  \\
MM-Trans~\cite{liu2022learning} & &  85.62 / 74.86  &  86.13 / 75.64  &  86.18 / 75.71  &  86.06 / 75.53  &  86.18 / 75.72  &  84.57 / 73.26  &  82.62 / 70.39  &  72.28 / 56.59  &  83.70 / 72.21 \\
SILI \cite{chen2023remote} &&87.01 / 77.01 & 87.65 / 78.02 & 88.21 / 78.90 & 88.55 / 79.44 & 88.38 / 79.18 & 86.73 / 76.57 & 84.31 / 72.87 & 73.13 / 57.64 & 85.50 / 74.95 \\ \hline
HiCD &$\downarrow+4\times$& \textbf{91.71 / 84.69}  &  \textbf{91.63 / 84.56}  &  \textbf{91.54 / 84.39}  &  \textbf{91.14 / 83.73}  &  \textbf{90.84 / 83.22}  &  \textbf{88.77 / 79.81}  &  86.11 / 75.61  &  66.61 / 49.93   & 87.29 / 78.24 \\
HiCD & degrdn.+$4\times$ & 91.21 / 83.85  &  91.19 / 83.80  &  91.09 / 83.64  &  90.56 / 82.75  &  90.15 / 82.07  &  88.28 / 79.01  &  \textbf{86.87 / 76.79}  &  83.34 / 71.44   & \textbf{89.09 / 80.42} \\
HiCD &degrdn.+$8\times$ &89.30 / 80.67 & 89.36 / 80.77 & 89.13 / 80.39 & 88.43 / 79.25 & 88.22 / 78.92 & 87.26 / 77.41 & 86.44 / 76.12 &  \textbf{84.95 / 73.84} & 87.89 / 78.42 \\ 
\hline
\end{tabular}
}
\label{tab:levir_res}
\vspace{3pt}
\begin{minipage}{\textwidth} 
\footnotesize
The results of comparative methods are directly taken from \cite{chen2023remote}. \textit{Data AUG.} is short for augmentation strategy of the training set. $\downarrow+4\times$ refers to the training set augmented by only downsampling with a ratio of 4, while degrdn.$+4\times$ and degrdn.$+8\times$ are the training set augmented by the degradation model with $4\times$ or $8\times$ downsampling. \textit{Avg.} is short for average performance. The best results are highlighted in \textbf{bold}.
\end{minipage}
\end{threeparttable}
\end{table*}

\begin{table*}[ht!]
\renewcommand{\arraystretch}{1.25}
\centering
\caption{Results (F1/IoU in \%) on the SV-CD dataset with the setting of only resolution difference.}
\begin{threeparttable}
\resizebox{1.0\textwidth}{!}{
\begin{tabular}{c|c|cccccccc|c}
\hline
\multirow{2}{*}{Method}& \multirow{2}{*}{Data AUG.}& \multicolumn{8}{c|}{ Different resolution difference ratio} & \multirow{2}{*}{Avg.} \\ \cline{3-10}
 & & $1\times$ & $2 \times$ & $4 \times$ & $5 \times$ & $8 \times$ & $9 \times$ & $10 \times$ & $12\times$ & \\
 \hline
 FC-EF \cite{daudt2018fully}  & \multirow{12}{*}{$\downarrow+8\times$}    & 55.88 / 38.77 & 55.88 / 38.77 & 55.96 / 38.85 & 55.99 / 38.88 & 56.06 / 38.95 & 56.06 / 38.94 & 56.06 / 38.95 & 56.03 / 38.92 & 56.01 / 38.89  \\
 FC-Siam-conc \cite{daudt2018fully}  & & 64.31 / 47.39 & 64.52 / 47.62 & 63.97 / 47.03 & 63.35 / 46.36 & 59.99 / 42.85 & 58.39 / 41.23 & 56.84 / 39.70 & 53.88 / 36.87  & 60.13 / 43.09  \\
 FC-Siam-diff \cite{daudt2018fully} &  & 67.35 / 50.77 & 67.41 / 50.84 & 67.09 / 50.48 & 66.86 / 50.22 & 64.97 / 48.12 & 63.54 / 46.56 & 61.84 / 44.76 & 58.51 / 41.35  & 64.32 / 47.48 \\
 STANet \cite{chen2020spatial} &   & 72.22 / 56.52 & 73.39 / 57.97 & 76.97 / 62.56 & 77.72 / 63.55 & 76.41 / 61.82 & 73.93 / 58.64 & 71.35 / 55.47 & 67.05 / 50.43 & 73.83 / 58.63 \\
 IFNet \cite{Zhang2020} &  & 81.54 / 68.83 & 81.97 / 69.45 & 84.17 / 72.66 & 85.34 / 74.42 & 86.61 / 76.38 & 86.34 / 75.97 & 85.69 / 74.97 & 83.77 / 72.07 & 84.84 / 73.70 \\

 SNUNet \cite{Fang2021}  & & 75.08 / 60.10 & 79.29 / 65.69 & 87.77 / 78.21 & 89.62 / 81.20 & 87.98 / 78.55 & 85.54 / 74.74 & 83.45 / 71.60 & 80.04 / 66.73 & 84.81 / 73.82 \\
 BIT \cite{BIT}  &  & 85.59 / 74.81 & 86.82 / 76.70 & 90.07 / 81.94 & 90.98 / 83.46 & 90.53 / 82.69 & 88.11 / 78.75 & 85.16 / 74.15 & 80.07 / 66.77 & 87.39 / 77.78  \\
 ICIFNet \cite{feng2022icif} &  &  91.20 / 83.83 & 91.56 / 84.44 & 92.83 / 86.63 & 93.25 / 87.35 & 93.05 / 87.00 & 91.95 / 85.10 & 90.65 / 82.90 & 88.02 / 78.60 & 91.62 / 84.57 \\
 DMINet \cite{Feng2023}  &  & 92.14 / 85.42 & 92.52 / 86.09 & 93.66 / 88.07 & 93.92 / 88.54 & 93.60 / 87.96 & 93.00 / 86.91 & 92.05 / 85.27 & 89.59 / 81.14  & 92.62 / 86.28\\ 

 SUNet \cite{Shao2021} &     &  67.12 / 50.51 & 67.33 / 50.76 & 69.88 / 53.70 & 72.44 / 56.80 & 77.10 / 62.73 & 77.55 / 63.33 & 77.69 / 63.52 & 76.90 / 62.46  & 74.13 / 59.04 \\
 SRCDNet \cite{Liu2022g}  &     & 78.14 / 64.13 & 82.07 / 69.59 & 89.19 / 80.49 & 90.67 / 82.93 & 91.59 / 84.49 & 90.76 / 83.08 & 89.29 / 80.66 & 85.19 / 74.19  & 88.39 / 79.35 \\
 MM-Trans~\cite{liu2022learning} & &  93.04 / 86.98  &  93.05 / 87.01  &  93.07 / 87.03  &  93.06 / 87.02 &  93.05 / 87.01  &  93.00 / 86.91  &  92.95 / 86.84  &  92.84 / 86.63  &  93.01 / 86.93  \\
 SILI \cite{chen2023remote} &     & 94.07 / 88.80 & 94.11 / 88.88 & 94.26 / 89.14 & 94.30 / 89.22 & 94.32 / 89.24 & 93.55 / 87.87 & 92.80/ 86.57 & 90.50 / 82.65  & 93.29 / 87.45  \\ \hline
HiCD & $\downarrow+8\times$ & \textbf{95.53 / 91.45}  &  \textbf{95.72 / 91.80}  &  \textbf{95.97 / 92.25}  &  \textbf{96.18 / 92.65}  &  \textbf{95.96 / 92.23}  &  \textbf{95.51 / 91.41}  &  94.95 / 90.38  &  93.34 / 87.50   & \textbf{95.40 / 91.21} \\
HiCD & degrdn.$+8\times$ &   95.31 / 91.03   &   95.45 / 91.29   &   95.61 / 91.58   &   95.65 / 91.67   &   95.46 / 91.31   &   95.27 / 90.98   &   \textbf{95.06 / 90.58}   &   \textbf{94.60 / 89.75} & 95.30 / 91.02   \\
\hline
\end{tabular}
}
\label{tab:ccd_res}
\vspace{3pt}
\begin{minipage}{\textwidth} 
The results of comparative methods are directly taken from \cite{chen2023remote}. \textit{Data AUG.} is short for augmentation strategy of the training set. $\downarrow+8\times$ refers to the training set augmented by only downsampling with a ratio of 8, while degrdn.$+8\times$ is the training set augmented by the degradation model with $8\times$ downsampling. \textit{Avg.} is short for average performance. The best results are highlighted in \textbf{bold}.     
\end{minipage}
\end{threeparttable}
\end{table*}

\begin{table*}[t!]
\renewcommand{\arraystretch}{1.25}
\centering
\caption{Results (F1/IoU in \%) on the BANDON dataset with the setting of only resolution difference.}
\begin{threeparttable}
\resizebox{\textwidth}{!}{
\begin{tabular}{c|c|cccccccc|c}
\hline
\multirow{2}{*}{Method} &\multirow{2}{*}{Data AUG.} & \multicolumn{8}{c|}{ Different resolution difference ratio } & \multirow{2}{*}{Avg.}  \\ \cline{3-10}
& &1× & 1.3× & 2× & 3× & 4× & 5× & 6× & 8× \\ \hline
ChangeFormer \cite{changeformer} & \multirow{2}{*}{Vanilla}&  67.63 / 51.09  &  67.52 / 50.97  &  54.18 / 37.16  &  66.34 / 49.63  &  51.08 / 34.30  &  58.77 / 41.61  &  41.43 / 26.13  &  31.69 / 18.83   & 54.83 / 38.71 \\

Baseline     &                       & 70.27 / 54.16  &  70.20 / 54.08  &  55.80 / 38.70  &  68.85 / 52.49  &  53.11 / 36.15  &  61.65 / 44.56  &  42.69 / 27.14  &  32.59 / 19.47   & 56.89 / 40.84 \\
\hline

FC-EF \cite{daudt2018fully} &\multirow{4}{*}{degrdn.+$8\times$}& 43.68 / 27.95  &  43.88 / 28.10  &  35.53 / 21.61  &  44.03 / 28.23  &  35.41 / 21.51  &  44.00 / 28.20  &  35.37 / 21.49  &  35.78 / 21.79   & 39.71 / 24.86 \\
FC-Siam-conc \cite{daudt2018fully} &                       &  51.93 / 35.08  &  52.97 / 36.03  &  42.42 / 26.92  &  54.25 / 37.23  &  42.91 / 27.32  &  53.52 / 36.54  &  41.31 / 26.03  &  39.03 / 24.25   & 47.29 / 31.17 \\

FC-Siam-diff \cite{daudt2018fully} &                       & 42.90 / 27.30  &  44.61 / 28.70  &  35.22 / 21.37  &  48.37 / 31.90  &  37.18 / 22.83  &  48.61 / 32.11  &  35.23 / 21.38  &  32.39 / 19.33   & 40.56 / 25.62 \\

ChangeFormer \cite{changeformer} &                       &  62.57 / 45.53  &  62.75 / 45.72  &  50.11 / 33.43  &  62.65 / 45.61  &  49.46 / 32.86  &  61.91 / 44.83  &  48.89 / 32.36  &  48.61 / 32.11   & 55.87 / 39.06 \\

HiCD         &                       & \textbf{69.50 / 53.26}  &  \textbf{69.54 / 53.31}  &  \textbf{54.76 / 37.70}  &  \textbf{68.96 / 52.63}  &  \textbf{53.72 / 36.72}  &  \textbf{67.82 / 51.30}  &  \textbf{52.35 / 35.45}  &  \textbf{51.59 / 34.76}   & \textbf{61.03 / 44.39} \\

\hline
\end{tabular}
}
\label{tab:bandon_res}

\vspace{3pt}
\begin{minipage}{\textwidth} 
\textit{Data AUG.} is short for augmentation strategy of the training set. \textit{Vanilla} indicates the training set augmented by random flip and random rotation, while degrdn.$+8\times$ refers to the training set augmented by the degradation model with $8\times$ downsampling. \textit{Avg.} is short for average performance. The best results are highlighted in \textbf{bold}.
\end{minipage}
\end{threeparttable}
\end{table*}

\begin{table*}[!t]
\renewcommand{\arraystretch}{1.25}
\centering
\caption{Results (F1/IoU in \%) on the LEVIR-CD and BANDON datasets with the setting of single-degradation quality difference.}
\vspace{-6pt}
\begin{threeparttable}

(a)  Results on LEVIR-CD dataset

\vspace{3pt}
\resizebox{\textwidth}{!}{
\begin{tabular}{c|c|c|cccc|cccc|cccc|c}
\hline
\multirow{2}{*}{Method} & \multirow{2}{*}{Data AUG.}
& \multirow{2}{*}{Clean} & \multicolumn{4}{|c|}{Blur} & \multicolumn{4}{|c|}{Noise}
& \multicolumn{4}{c|}{Digital} & \multirow{2}{*}{Avg.} \\ \cline{4-15}

& & & Motion & Defocus & Glass & Gaussian & Gaussian & Impulse & Shot & Mixed & Brightness & Pixelate & Saturation & JPEG & \\ 
\hline
ChangeFormer \cite{changeformer} & \multirow{2}{*}{Vanilla} & 90.79 / 83.13 & 46.60 / 37.22 & 33.92 / 26.18 & 43.77 / 34.78 & 42.16 / 34.21 & 57.05 / 46.43 & 61.13 / 48.33 & 64.87 / 52.75 & \textbf{83.59 / 72.49} & 89.05 / 80.30 & 88.18 / 78.92 & 78.66 / 67.21 & 77.29 / 64.98 & 63.86 / 53.65   \\
Baseline & & \textbf{91.63 / 84.55} & 50.91 / 41.12 & 36.02 / 27.68 & 47.46 / 37.83 & 42.44 / 34.34 & 51.42 / 40.99 & 55.38 / 43.79 & 56.96 / 45.68 & 80.16 / 68.50 & \textbf{90.58 / 82.80} & \textbf{89.27 / 80.67} & 80.44 / 69.80 & 81.72 / 69.99 & 63.56 / 53.60  \\
\hline
FC-EF \cite{daudt2018fully} & \multirow{5}{*}{degrdn.+$8\times$} &76.87 / 62.43& 67.94 / 52.04 & 68.99 / 53.78 & 75.61 / 60.91 & 67.38 / 52.79 & 69.78 / 53.86 & 70.51 / 54.58 & 69.24 / 53.28 & 73.02 / 57.65 & 39.14 / 26.63 & 77.86 / 63.75 & 41.51 / 28.88 & 75.55 / 60.73 &  66.38 / 51.57 \\
FC-Siam-conc \cite{daudt2018fully} & &  60.69 / 43.56 & 60.34 / 45.12 & 72.03 / 57.34 & 76.55 / 62.08 & 70.46 / 56.43 & 17.24 / 10.01 & 13.46 / 7.77 & 14.14 / 8.21 & 23.22 / 14.02 & 46.19 / 31.48 & 67.46 / 50.94 & 27.53 / 17.29 & 62.88 / 45.91 & 45.96 / 33.88  \\
FC-Siam-diff \cite{daudt2018fully} \cite{daudt2018fully} & & 60.37 / 43.23 & 61.89 / 46.45 & 71.48 / 56.25 & 75.87 / 61.13 & 67.38 / 52.90 & 18.61 / 10.83 & 20.03 / 12.07 & 18.52 / 10.96 & 29.76 / 18.59 & 47.68 / 32.57 & 64.33 / 47.43 & 28.66 / 18.08 & 61.81 / 44.78 & 47.17 / 34.34 \\
ChangeFormer \cite{changeformer} & & 83.40 / 71.53 & \textbf{75.18 / 60.64} & 80.73 / 68.08 & 84.10 / 72.59 & 78.65 / 66.00 & \textbf{79.17 / 65.62} & 79.94 / 66.62 & \textbf{79.70 / 66.30} & 81.29 / 68.50 & 76.30 / 61.94 & 84.45 / 73.08 & 73.46 / 58.98 & 80.22 / 67.05 & 79.43 / 66.28  \\
HiCD & & 89.30 / 80.67 & 60.94 / 48.63 & \textbf{84.30 / 73.16} & \textbf{85.36 / 74.57} & \textbf{84.68 / 73.85} & 73.22 / 59.19 & \textbf{80.77 / 69.13} & 76.12 / 62.31 & 83.52 / 71.97 & 86.70 / 76.58 & 88.53 / 79.42 & \textbf{82.10 / 70.51} & \textbf{84.01 / 72.57} & \textbf{81.42 / 70.01} \\

\hline
\end{tabular}
}
\label{tab:levir_single_deg}

(b)  Results on BANDON dataset

\vspace{3pt}
\resizebox{\textwidth}{!}{
\begin{tabular}{c|c|c|cccc|cccc|cccc|c}
\hline
\multirow{2}{*}{Method} & \multirow{2}{*}{Data AUG.}
& \multirow{2}{*}{Clean} & \multicolumn{4}{|c|}{Blur} & \multicolumn{4}{|c}{Noise}
& \multicolumn{4}{|c|}{Digital}  & \multirow{2}{*}{Avg.} \\ \cline{4-15}

& & & Motion & Defocus & Glass & Gaussian & Gaussian & Impulse & Shot & Mixed & Brightness & Pixelate & Saturation & JPEG \\ 
\hline
ChangeFormer \cite{changeformer} & \multirow{2}{*}{Vanilla} & 67.63 / 51.09 & 38.79 / 24.52 & 37.34 / 23.66 & 33.14 / 21.00 & 39.80 / 25.62 & 41.78 / 26.57 & 45.81 / 29.83 & 40.09 / 25.24 & 44.47 / 28.75 & 48.92 / 32.63 & 52.93 / 36.00 & \textbf{49.36 / 32.91} & 49.83 / 33.25 & 43.52 / 28.33 \\

Baseline & & \textbf{70.27 / 54.16}  & 40.05 / 25.52 & 39.70 / 25.38 & 35.07 / 22.48 & 41.20 / 26.74 & 42.80 / 27.48 & 46.63 / 30.56 & 40.58 / 25.74 & 45.23 / 29.47 & \textbf{49.73 / 33.44} & \textbf{54.69 / 37.65} & 48.78 / 32.60 & 50.74 / 34.07 & 44.60 / 29.26 \\
\hline
FC-EF \cite{daudt2018fully} & \multirow{5}{*}{degrdn.+$8\times$} & 43.68 / 27.95 & 34.29 / 20.70 & 34.10 / 20.56 & 35.11 / 21.29 & 34.02 / 20.51 & 25.49 / 14.61 & 26.65 / 15.38 & 24.06 / 13.69 & 26.17 / 15.10 & 17.09 / 9.65 & 35.55 / 21.62 & 23.03 / 13.25 & 34.54 / 20.88 & 29.18 / 17.27 \\

FC-Siam-conc \cite{daudt2018fully} & & 51.93 / 35.08 & 37.40 / 23.06 & 38.84 / 24.19 & 36.31 / 22.34 & 39.23 / 24.49 & 20.07 / 11.20 & 20.29 / 11.32 & 18.03 / 9.94 & 19.95 / 11.12 & 27.33 / 16.11 & 38.03 / 23.49 & 33.80 / 20.42 & 36.54 / 22.37 & 30.49 / 18.34 \\

FC-Siam-diff \cite{daudt2018fully} & & 42.90 / 27.30  & 31.13 / 18.46 & 33.69 / 20.30 & 31.96 / 19.09 & 33.65 / 20.28 & 18.86 / 10.41 & 17.81 / 9.77 & 18.71 / 10.32 & 18.36 / 10.11 & 25.62 / 14.78 & 28.62 / 16.71 & 27.77 / 16.13 & 27.49 / 15.94 & 26.14 / 15.19 \\

ChangeFormer \cite{changeformer} & & 62.57 / 45.53 & 45.19 / 29.27 & 47.15 / 30.89 & 47.54 / 31.21 & 47.26 / 31.00 & 42.77 / 27.23 & 45.02 / 29.08 & 41.98 / 26.61 & 44.08 / 28.31 & 40.47 / 25.66 & 49.64 / 33.02 & 44.33 / 28.66 & 47.91 / 31.52 & 45.28 / 29.37 \\

HiCD & &  69.50 / 53.26 & \textbf{46.93 / 30.82} & \textbf{50.12 / 33.52} & \textbf{47.81 / 31.61} & \textbf{50.71 / 34.06} & \textbf{48.68 / 32.20} & \textbf{49.52 / 32.94} & \textbf{47.91 / 31.54} & \textbf{49.67 / 33.07} & 47.90 / 31.84 & 54.01 / 37.00 & 46.77 / 30.96 & \textbf{51.41 / 34.64} & \textbf{49.29 / 32.85} \\
\hline
\end{tabular}
}\label{tab:bandon_single_deg}

\vspace{3pt}
\begin{minipage}{\textwidth} 
\textit{Data AUG.} is short for augmentation strategy of the training set. \textit{Vanilla} indicates the training set augmented by random flip and random rotation, while degrdn.$+8\times$ refers to the training set augmented by the degradation model with $8\times$ downsampling. \textit{Avg.} and \textit{JPEG} are short for average performance and JPEG compression, respectively. \textit{Clean} refers to the original testing set without any degradation. The best results are highlighted in \textbf{bold}.
\end{minipage}
\end{threeparttable}
\end{table*}

\textbf{Only Resolution Difference.} In this type of experiment, we compare the proposed HiCD with excellent methods under two kinds of training sets. The first training set follows what is done in~\cite{chen2023remote} for resolution-difference change detection: each training set is augmented by only downsampling with the fixed ratio, such as $4\times$ and $8\times$. The second training set is augmented by the degradation model in Equation~\eqref{degradation-model}. Results on three datasets are reported in Table~\ref{tab:levir_res}, Table~\ref{tab:ccd_res} and Table~\ref{tab:bandon_res}, respectively. Specifically,

\textit{Results on LEVIR-CD dataset.} With increasing resolution differences between bi-temporal images, models trained on LEVIR-CD ($\downarrow$+$4\times$) suffer significant performance drops. In the most challenging scenario ($8\times$), existing models except for FC-EF, SUNet, MM-Trans and SILI, reach IoU inferior to 50\%. The best among the four is SILI, which obtains 57.64\% on the 8$\times$ scenario, and 74.95\% on average IoU. In terms of average performance, HiCD outperforms existing models. In particular, HiCD trained on LEVIR-CD ($\downarrow$+$4\times$), aka., HiCD-$\downarrow$-$4\times$ gains \textbf{3.29\%} IoU, HiCD-degrdn.-$4\times$ \textbf{5.47\%} and HiCD-degrdn.-$8\times$ \textbf{3.47\%}. Compared to HiCD-$\downarrow$-$4\times$,  HiCD-degrdn.-$4\times$ and  HiCD-degrdn.-$8\times$ surpass other models by a large margin on the 8$\times$ scenario. The average improvement is \textbf{15\%}. Thus, we argue that the CD model is exposed to hard samples caused by diverse degradation during training, enabling the model to better deal with continuous resolution differences.

\textit{Results on SV-CD dataset.} Among eight scenarios, most general change detection methods perform worst on the 12$\times$ scenario. However, resolution-difference change detection methods SUNet and SRCDNet get the worst result on 1$\times$ scenario. Specifically, SUNet obtains average IoU 63.07\% across 8$\times$ to 12$\times$ scenarios, and 52.9\% across 1$\times$ to 5$\times$ scenarios. The highest performance gap between different scenarios (50.51\% on 1$\times$ vs 63.52\% on 10$\times$) is more than 13\%. This demonstrates that SUNet achieves good results on quality-varied image pairs, at the cost of the ability to deal with equal and high-quality pairs. SRCDNet surpasses SUNet in all scenarios by a large margin, achieving an average improvement of more than 20\%. However, it still suffers a significant performance drop on the 1$\times$ scenario. Compared to the above methods, MM-Trans and SILI are able to maintain good performance on each scenario while obtaining a further boost. MM-Trans gains 7.58\% on average IoU, and SILI 8.1\%. Overall, the best method is HiCD, exceeding other methods by more than 3.5\% on average IoU. Moreover, HiCD achieves performance superior to \textbf{90\%} on the first seven scenarios, regardless of data augmentation strategies. For the challenging 12 $\times$, HiCD-degrdn.-$8\times$ obtains \textbf{89.75\%} IoU, which is \textbf{3.12\%} better than the current SOTA one reported by MM-Trans. These results demonstrate the superiority of our method.

\textit{Results on BANDON dataset.} The BANDON dataset is a challenging building change detection dataset composed of off-nadir aerial images. The baseline and excellent ChangeFormer trained on the original training set (\emph{i.e.}, dataset without downsampling or degradation) can only obtain 54.16\% and 51.09\% IoU on the original testing set, respectively. Furthermore, they suffer a large performance drop when bi-temporal resolution differences increase. General change detection models may be adapted to the resolution-difference change detection via training on the BANDON (degrdn.+$8\times$) dataset with the cross-entropy loss, but at the cost of a performance drop on the original testing set. For example, ChangeFormer-degrdn.-$8\times$ gains 13.28\% IoU in the $8\times$ case, but losses 5.56\% IoU in the $1\times$. Compared to general CD models trained on BANDON (degrdn.+$8\times$), HiCD-degrdn.-$8\times$ not only improves performances on the resolution-difference change detection, but also obtains performances closer to the baseline on the equal-resolution change detection. This confirms the effectiveness of the knowledge distillation-based training strategy to transfer the teacher's deep priors about change detection.

\textbf{Single-degradation Quality Difference.} We conduct this type of experiment to evaluate our method for handling quality differences that are caused by other individual degradations except for resolution downsampling. The chosen degradation covers blur, noise, and digital categories. For blur and digital categories, each category has four kinds of variants, each with five levels of severity. Following~\cite{xie2021segformer}, four types of the noise category only involve the first three levels of severity. Consequently, there are 52 distinct degradations. For quantitative evaluation, we calculate performances on each variant by averaging performances on its five or three severity levels. Besides, we average performances on 12 variants to get overall performance. Results are reported in Tables~\ref{tab:levir_single_deg}.

\begin{figure*}[htb!]
    \centering \includegraphics[width=0.9\linewidth]{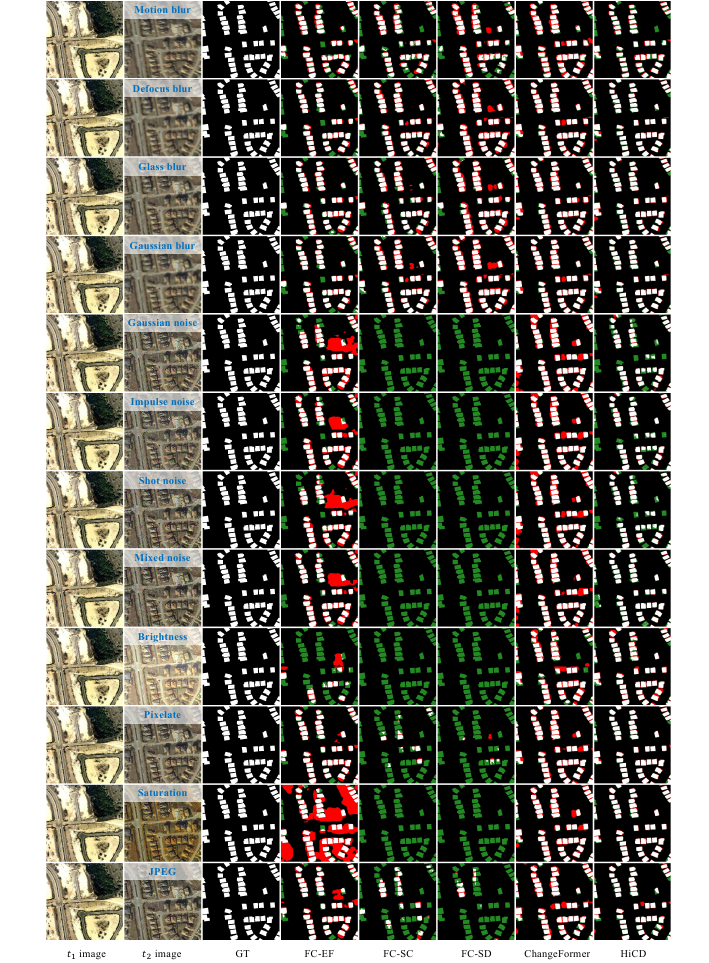}
    \caption{Qualitative results on the LEVIR-CD dataset with the setting of single-degradation quality difference. \textit{GT}, \textit{FC-SC}, and \textit{FC-SD} are short for the ground truth, FC-Siam-conc \cite{daudt2018fully} and FC-Siam-diff \cite{daudt2018fully}, respectively. Different colors, \emph{i.e.}, white, black, red and green, indicate true positive, true negative, false positive and false negative.}
    \label{fig: single_deg_resluts}
\end{figure*}

\begin{figure*}[ht!]
    \centering
    \includegraphics[width=0.95\textwidth]{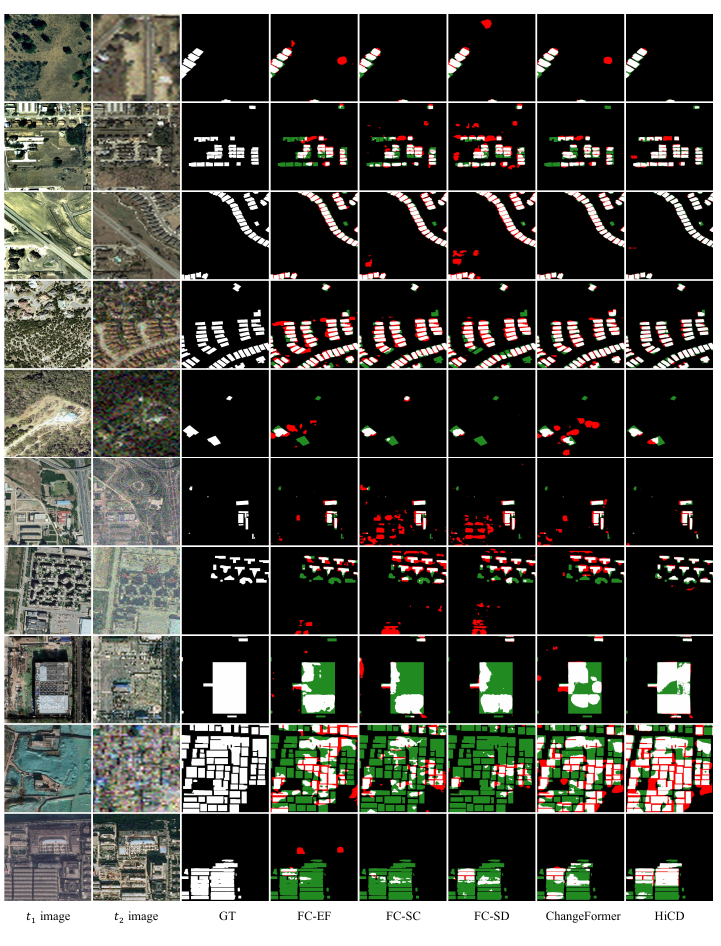}
    \caption{Qualitative results on the LEVIR-CD and BANDON datasets with the setting of multi-degradation quality difference. \textit{GT}, \textit{FC-SC}, and \textit{FC-SD} are short for the ground truth, FC-Siam-conc \cite{daudt2018fully} and FC-Siam-diff \cite{daudt2018fully}, respectively. Different colors, \emph{i.e.}, white, black, red and green, indicate true positive, true negative, false positive and false negative.}
    \label{fig: multi_deg_resluts}
\end{figure*}

\begin{figure*}[!t]
    \centering
    \includegraphics[width=0.95\textwidth]{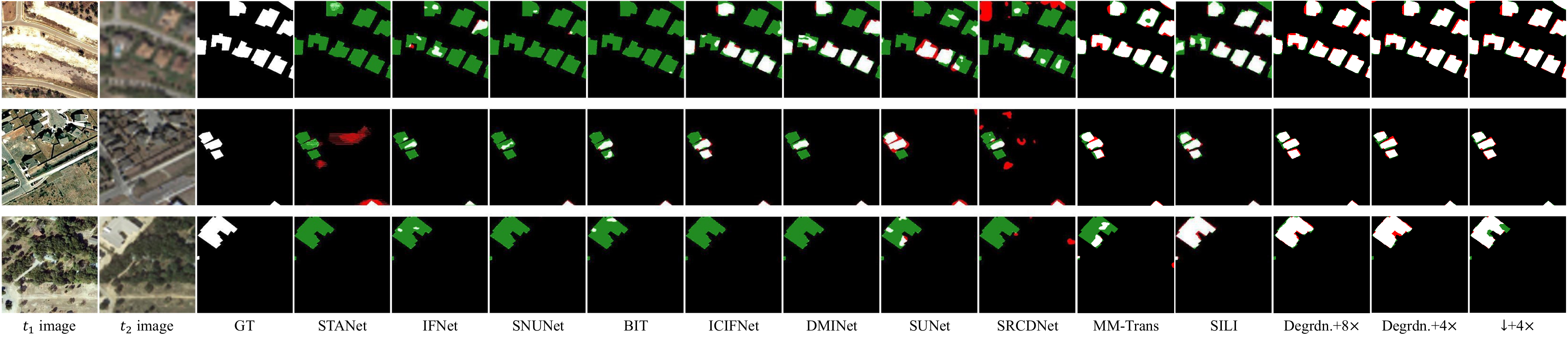}
    \caption{Qualitative results on the LEVIR-CD dataset with the setting of 4$\times$ resolution difference. \emph{GT} is short for the ground truth. Different colors, \emph{i.e.}, white, black, red and green, indicate true positive, true negative, false positive and false negative.}
    \label{fig:sili_resolution}
\end{figure*}

\textit{Results on LEVIR-CD dataset.} With LEVIR-CD (degrdn.+$8\times$) dataset, Transformer-based ChangeFormer surpasses CNN-based FC-EF, FC-Siam-conc and FC-Siam-diff in both clean (6.65\% in F1 and 9.1\% in IoU) and quality-varied (13.05\% in average F1 and 14.71\% in average IoU) testing set. Despite that, HiCD-degrdn.-$8\times$ can obtain a further boost, where we gain \textbf{9.14\%} IoU and \textbf{3.73\%} average IoU for clean and quality-varied cases, respectively. It is especially interesting to compare HiCD-degrdn.-$8\times$ and ChangeFormer with their baseline which is trained on the original training set (\emph{i.e.}, dataset without degradation). ChangeFormer, Compared to its baseline, gains 15.57\% average F1 and 12.65\% average IoU for quality-varied cases, but reduces 7.39\% F1 and 11.6\% IoU for the clean case. However, HiCD-degrdn.-$8\times$ is not only far superior to its baseline for quality-varied cases (\textbf{17.86\%} in average F1 and \textbf{16.42\%} in average IoU), but also close to the baseline for the clean case where the gap is 2.33\% in F1 and 3.88\% in IoU. This demonstrates the potential of the novel training strategy to facilitate real applications of change detection models.

\textit{Results on BANDON dataset.} General change detection models trained on BANDON (degrdn.+$8\times$) achieve poor trade-offs between dealing with equal high-quality image pairs and quality-varied image pairs. ChangeFormer obtains an improvement for quality-varied cases, at the cost of a performance drop for the clean cases. Again,  HiCD-degrdn.-$8\times$ achieves the best trade-off, outperforming the baseline by 3.59\% average IoU for quality-varied cases and reaching a slightly lower IoU 53.26\% for the clean case. Remarkably, HiCD-degrdn.-$8\times$ gets the highest F1/IoU on nine degradations (\emph{i.e.}, four types of blur, four types of noise, and JPEG compression). The highest gap (5.93\% in F1 and 4.94\% in IoU) is on the \emph{shot noise} degradation. However, it is slightly inferior to its baseline on degradations of brightness, pixelate, and saturation. One possible reason is that the training set is augmented by the degradation model consisting of Gaussian blur, Gaussian noise, and downsampling. Since the CD model has been exposed to quality differences caused by similar degradations during training, HiCD achieves good results on the blur and noise categories. Unlike blur, which results in the loss of fine details, and noise, which introduces random variations in pixel values, the degradation of digital categories usually leads to color changes in the image. Thus, HiCD can only maintain a performance similar to its baseline. This issue can be addressed by designing a more complex yet practical degradation model.

\textbf{Multi-degradation Quality Difference.} The drop in image quality can be caused by a combination of multiple degradations. Hence, we conduct this type of experiment to further verify the capability of HiCD. Particularly, we utilize the degradation model with 4$\times$ and 8$\times$ downsampling to process the LEVIR-CD and BANDON datasets respectively, creating the corresponding testing sets. From Table~\ref{tab_levir_deg}, we can observe that the baseline and ChangeFormer trained on original datasets work poorly for this experimental setting, reaching IoU inferior to 2\% on the LEVIR-CD dataset and 14\% on the BANDON dataset. Training with datasets augmented by the degradation model in Equation~\eqref{degradation-model} provides benefits. Compared to ChangeFormer-vanilla, ChangeFormer-degrdn.-$8\times$ improves the IoU of 63.8\% on LEVIR-CD and 29.16\% on BANDON. Despite that, HiCD-degrdn.-$8\times$ significantly exceeds ChangeFormer by 9.6\% and 4.33\% on the above datasets, respectively.

\textbf{Visualization.} We report some qualitative results to show the superiority of our method intuitively. Different colors are used to represent true positive (white), true negative (black), false positive (red) and false negative (green). From Figure~\ref{fig: resolution_resluts}, we can observe that the simple interpolation operation cannot recover object details that are damaged or lost due to downsampling, posing great challenges to change detection. Existing CD methods yield many false negatives. However, HiCD achieves consistent accurate detection regardless of resolution difference ratios. Following~\cite{chen2023remote}, we show more results of comparison methods, as shown in Figure~\ref{fig:sili_resolution}. Methods except for SILI fail to identify building changes. While SILI outperforms other methods, HiCD is able to produce more complete detection results.

Figure~\ref{fig: single_deg_resluts} shows visual comparisons of single-degradation quality difference. FC-Siam-diff and FC-Siam-conc fail to detect the change of interest when the bi-temporal quality difference is caused by noise and digital categories. FC-EF and ChangeFormer outperform them but introduce many false positives. The results of the proposed HiCD are closest to the ground truth. In Figure~\ref{fig: multi_deg_resluts}, we report the comparisons of multi-degradation quality difference. The combination of blur, noise and downsampling leads to very poor visual quality of $t_2$ images. It is difficult for humans to recognize the changes, especially the ninth image pair. However, HiCD still is able to detect most building changes. All the visual comparisons demonstrate that our HiCD can handle different quality-varied challenges and yield accurate change detection results.

\begin{table}[ht!]
\renewcommand{\arraystretch}{1.25}
\centering
\fontsize{5}{5.8}\selectfont 
\caption{Results (F1/IoU in \%) on the LEVIR-CD and BANDON dataset with the setting of multi-degradation quality difference.}
\begin{threeparttable}
\resizebox{\linewidth}{!}{
\begin{tabular}{c|c|cc}

\hline
Method & Data AUG. & LEVIR-CD & BANDON \\
\hline
ChangeFormer \cite{changeformer} & \multirow{2}{*}{Vanilla}    & 2.56 / 1.30 & 21.82 / 12.25 \\
Baseline     &                           & 3.07 / 1.56 & 24.20 / 13.77  \\
\hline
FC-EF \cite{daudt2018fully}       & \multirow{5}{*}{degrdn.+$8\times$} & 73.54 / 58.15 & 42.62 / 27.08  \\
FC-Siam-conc \cite{daudt2018fully} &                           & 71.43 / 55.56 & 45.07 / 29.09   \\
FC-Siam-diff \cite{daudt2018fully} &                           & 69.97 / 53.82 & 40.46 / 25.36   \\
ChangeFormer \cite{changeformer} &                           & 78.86 / 65.10 & 58.57 / 41.41    \\
HiCD         &                           & \textbf{85.52 / 74.70} & \textbf{62.77 / 45.74}  \\

\hline
\end{tabular}
}
\label{tab_levir_deg}

\vspace{3pt}
\fontsize{7.5}{8.5}\selectfont 
\begin{minipage}{\linewidth} 
\textit{Data AUG.} is short for augmentation strategy of the training set. \textit{Vanilla} indicates the training set augmented by random flip and random rotation, while degrdn.$+8\times$ refers to the training set augmented by the degradation model with $8\times$ downsampling. The best results are highlighted in \textbf{bold}.
\end{minipage}
\end{threeparttable}
\end{table}

\subsection{Ablation Studies}
We perform ablation experiments to verify the effect of the details of our method. Specifically,

\textbf{Different teacher networks.}
We compare our baseline with ChangeFormer on the LEVIR-CD and BANDON datasets, to choose a strong teacher network. Because ChangeFormer is a recent popular work on general change detection and has been utilized by most of the previous related works. From Table~\ref{tab:levir_single_deg}~(a) and Table~\ref{tab:bandon_single_deg}~(b), we can observe that for general change detection task, our baseline outperforms ChangeFormer on both datasets. In terms of IoU, we gain 1.42\% on LEVIR-CD and 3.07\% on BANDON. Thus, we choose the baseline's architecture to construct the teacher network since it can capture significant deep priors about change detection.

\textbf{Effectiveness of hierarchical correlation distillation.}
Under the setting of multi-degradation quality difference, we validate the effect of the presented hierarchical correlation distillation by only performing semantic feature distillation with different correlation transfers. Results are reported in Table~\ref{tab:abl_1}. We start from the baseline which is optimized only with cross-entropy loss. We first add to the baseline self-correlation distillation loss ($\mathcal{L}_\mathrm{s1}$ and $\mathcal{L}_\mathrm{s2}$). The improvement is remarkable in terms of IoU, where we gain 4.26\% with $\mathcal{L}_\mathrm{s2}$ and 8.84\% with two self-correlation losses. Second, we add the cross-correlation distillation loss ($\mathcal{L}_\mathrm{c}$) to the semantic feature distillation module. Our $\mathcal{L}_\mathrm{c}$ provides a boost on the performance, which reaches 74.22\% IoU and 85.20\% F1. Finally, we add the global-correlation distillation loss ($\mathcal{L}_\mathrm{g}$). This component provides a slight improvement, where we gain 0.28\% in IoU and 0.18\% in F1. Though the improvement is small, it does not mean that our $\mathcal{L}_\mathrm{g}$ is dispensable. We additionally train the baseline with cross-entropy loss and global-correlation distillation loss. It can be observed that $\mathcal{L}_\mathrm{g}$ provides an improvement of about 8\% in IoU. This confirms the importance of accounting for correlation information across the whole training dataset to facilitate the learning of the student. Since self-correlation, cross-correlation and global-correlation are complementary, the baseline with hierarchical correlation distillation loss achieves the best performance.

\begin{table}[t!]
\renewcommand{\arraystretch}{1.25}
\centering
\caption{Ablation studies on the LEVIR-CD dataset with the setting of multi-degradation quality difference.}
\begin{threeparttable}
\resizebox{\linewidth}{!}{
\begin{tabular}{c c c c c c |c c  c c}
\hline
Baseline & $\mathcal{L}_\mathrm{s2}$  &  $\mathcal{L}_\mathrm{s1}$ & $\mathcal{L}_\mathrm{c}$ & $\mathcal{L}_\mathrm{g}$ & $\mathcal{L}_\mathrm{cfd}$ & IoU & F1 & Recall & Precision \\
\hline
$\checkmark$&&&&&&64.90 & 78.71 & 72.84 & 85.62 \\
$\checkmark$&$\checkmark$&&&&&69.16 & 81.77 & 78.64 & 85.17 \\
$\checkmark$&$\checkmark$&$\checkmark$&&&&73.74 & 84.88 & 83.40 & 86.43 \\
$\checkmark$&$\checkmark$&$\checkmark$&$\checkmark$&&&74.22 & 85.20 & \textbf{83.88} & 86.56 \\
$\checkmark$&$\checkmark$&$\checkmark$&$\checkmark$&$\checkmark$&&74.50 & 85.38 & 83.77 & 87.06 \\
$\checkmark$&&&&$\checkmark$&&72.95 & 84.36 & 81.54 & 87.38 \\
$\checkmark$&&&&&$\checkmark$&73.15 & 84.50 & 82.49 & 86.60 \\
$\checkmark$&$\checkmark$&$\checkmark$&$\checkmark$&$\checkmark$&$\checkmark$&\textbf{74.70} & \textbf{85.52} & 83.58 & \textbf{87.55} \\

\hline
\end{tabular}
}
\label{tab:abl_1}

\vspace{3pt}
\begin{minipage}{\linewidth} 
\textit{Baseline} is the student model trained only with cross-entropy loss. $\mathcal{L}_\mathrm{s1}$, $\mathcal{L}_\mathrm{s2}$, $\mathcal{L}_\mathrm{c}$ and $\mathcal{L}_\mathrm{g}$ denote hierarchical correlation distillation in the semantic feature distillation module. $\mathcal{L}_\mathrm{cfd}$ is the distillation loss of the change feature distillation module. The best results are highlighted in \textbf{bold}.
\end{minipage}
\end{threeparttable}
\end{table}

\textbf{Effecitveness of semantic and change feature distillation.}
We report a detailed analysis of the proposed distillation modules, considering the setting of multi-degradation quality difference. As shown in Table~\ref{tab:abl_1}, compared to the baseline, $\mathcal{L}_\mathrm{sdf}$ provides an improvement of 9.6\% in IoU, and $\mathcal{L}_\mathrm{cdf}$ improves IoU of 8.25\%. This shows that the student‘s learning can greatly benefit from the deep priors acquired by the teacher. Notice that the improvement of $\mathcal{L}_\mathrm{sdf}$ is larger than that of $\mathcal{L}_\mathrm{cdf}$. We argue that two losses provide mutual benefits. Thus, we use them together with cross-entropy loss by default, and achieve the best performance of 74.7\% IoU and 85.52\% F1 on the LEVIR-CD dataset.

\begin{table}[!t]
    \renewcommand{\arraystretch}{1.25}
    \centering
    \caption{Evaluation of $\lambda_\text{sfd}$ and $\lambda_\text{cfd}$ on the LEVIR-CD dataset with the setting of multi-degradation quality difference.}
    \begin{threeparttable}
    \begin{tabular}{ll|cccc}
    \hline
    \multicolumn{2}{c|}{Weight setting} & IoU & F1 & Recall & Precision \\
    \hline
    $\lambda_\text{sfd}=25$ & \multirow{5}{*}{$\lambda_\text{cfd}=0.25$}& 75.34 &85.93 &84.88 &87.01  \\
    $\lambda_\text{sfd}=10$ &&75.49 &86.03 &84.96 &87.13    \\
    $\lambda_\text{sfd}=5$ & &74.70 & 85.52	& 83.58 &87.55  \\
    $\lambda_\text{sfd}=2.5$  && 75.40 &85.98& 85.07 &86.90  \\
    $\lambda_\text{sfd}=1$ & & 74.33& 85.27& 83.25 &87.40 \\
    \hline
    \multirow{5}{*}{$\lambda_\text{sfd}=5$} & $\lambda_\text{cfd}=1.25$ &75.57 &86.09& 84.87 & 87.33 \\
    & $\lambda_\text{cfd}=0.5$ &74.70 & 85.52	& 83.58 &87.55  \\
    & $\lambda_\text{cfd}=0.25$ &75.24 &85.87& 84.20& 87.60  \\
    & $\lambda_\text{cfd}=0.125$ &75.46& 86.01 &85.48& 86.56 \\
    & $\lambda_\text{cfd}=0.05$ &75.32& 85.92& 84.66 &87.23\\
    \hline
    \end{tabular}
    \label{tab-weight}
    \vspace{3pt}
    \begin{minipage}{\linewidth} 
    $\mathcal{L}_\mathrm{sfd}$, $\mathcal{L}_\mathrm{cfd}$ are distillation losses of the semantic feature distillation module and change feature distillation module, respectively. For all experiments in this article, $\mathcal{L}_\mathrm{sfd}$ equals 5 and $\mathcal{L}_\mathrm{cfd}$ is 0.25.
    \end{minipage}
    \end{threeparttable}
\end{table}

\textbf{Parameter sensitivity.} As the key idea of our method is to leverage task knowledge to guide the representation learning and feature alignment of the CD model on quality-varied image pairs, we conduct experiments to evaluate the impact of loss weights of the semantic feature distillation module and the change feature distillation module, \emph{i.e.}, $\lambda_\text{sfd}$ and $\lambda_\text{cfd}$. As shown in Table~\ref{tab-weight}, our model gets the best results (75.57\% IoU) when $\lambda_\text{sfd}$ equals 5 and $\lambda_\text{cfd}$ is 1.25. In the case of rough selection of hyperparameters ($\lambda_\text{sfd}=5$ and $\lambda_\text{cfd}=0.25$), HiCD has produced promising results, thereby demonstrating the effectiveness of the proposed training strategy. Furthermore, we note that the performance of our model is around 75\% for ten hyperparameter settings. This illustrates that our method is not highly sensitive to the settings of these two parameters.

\section{Conclusion}
\label{Conclusion}
In this article, we focus on change detection with bi-temporal quality differences caused by diverse degradations, and propose a knowledge distillation-based training strategy to handle the quality-varied challenge. To effectively transfer the teacher's deep priors about change detection tasks, we propose hierarchical correlation distillation, \emph{i.e.}, self-correlation, cross-correlation and global-correlation distillations. On this basis, the semantic and change feature distillation modules are presented to improve the student's capability to extract good feature representation from low-quality images, and mine essential change features from quality-varied representations. Extensive experiments on commonly used change detection datasets demonstrate the superiority of our method. Our HiCD significantly outperforms recent excellent change detection methods in experimental settings of only resolution difference, single-degradation and multi-degradation quality differences. Besides, we conduct ablation experiments to further verify the effectiveness of the details of HiCD. Future works might consider the application of hierarchical correlation distillation to new tasks, such as incremental learning of change detection, or addressing the issue of quality difference in image pairs with different modalities~\cite{zhang2016change}.



\ifCLASSOPTIONcaptionsoff
  \newpage
\fi

\FloatBarrier
\bibliographystyle{ieeetr}
\bibliography{main}

\begin{thebibliography}{10}

\bibitem{howarth1983landsat}
P.~J. Howarth and E.~Boasson, ``Landsat digital enhancements for change
  detection in urban environments,'' {\em Remote sensing of environment},
  vol.~13, no.~2, pp.~149--160, 1983.

\bibitem{viana2019land}
C.~M. Viana, S.~Oliveira, S.~C. Oliveira, and J.~Rocha, ``Land use/land cover
  change detection and urban sprawl analysis,'' in {\em Spatial modeling in GIS
  and R for earth and environmental sciences}, pp.~621--651, Elsevier, 2019.

\bibitem{hegazy2015monitoring}
I.~R. Hegazy and M.~R. Kaloop, ``Monitoring urban growth and land use change
  detection with gis and remote sensing techniques in daqahlia governorate
  egypt,'' {\em International Journal of Sustainable Built Environment},
  vol.~4, no.~1, pp.~117--124, 2015.

\bibitem{zheng2021building}
Z.~Zheng, Y.~Zhong, J.~Wang, A.~Ma, and L.~Zhang, ``Building damage assessment
  for rapid disaster response with a deep object-based semantic change
  detection framework: From natural disasters to man-made disasters,'' {\em
  Remote Sensing of Environment}, vol.~265, p.~112636, 2021.

\bibitem{washaya2018coherence}
P.~Washaya, T.~Balz, and B.~Mohamadi, ``Coherence change-detection with
  sentinel-1 for natural and anthropogenic disaster monitoring in urban
  areas,'' {\em Remote Sensing}, vol.~10, no.~7, p.~1026, 2018.

\bibitem{trinder2012aerial}
J.~Trinder and M.~Salah, ``Aerial images and lidar data fusion for disaster
  change detection,'' {\em ISPRS Annals of the Photogrammetry, Remote Sensing
  and Spatial Information Sciences}, vol.~1, pp.~227--232, 2012.

\bibitem{sophiayati2009onboard}
S.~Sophiayati~Yuhaniz and T.~Vladimirova, ``An onboard automatic change
  detection system for disaster monitoring,'' {\em International Journal of
  Remote Sensing}, vol.~30, no.~23, pp.~6121--6139, 2009.

\bibitem{daudt2018fully}
R.~C. Daudt, B.~{Le Saux}, and A.~Boulch, ``{Fully convolutional siamese
  networks for change detection},'' in {\em ICIP}, pp.~4063--4067, 2018.

\bibitem{liu2020building}
Y.~Liu, C.~Pang, Z.~Zhan, X.~Zhang, and X.~Yang, ``{Building Change Detection
  for Remote Sensing Images Using a Dual-Task Constrained Deep Siamese
  Convolutional Network Model},'' {\em IEEE GRSL}, vol.~18, no.~5,
  pp.~811--815, 2020.

\bibitem{changeformer}
W.~G.~C. Bandara and V.~M. Patel, ``{A transformer-based siamese network for
  change detection},'' in {\em IGARSS 2022-2022 IEEE International Geoscience
  and Remote Sensing Symposium}, pp.~207--210, IEEE, 2022.

\bibitem{BIT}
H.~Chen, Z.~Qi, and Z.~Shi, ``{Remote sensing image change detection with
  transformers},'' {\em IEEE Transactions on Geoscience and Remote Sensing},
  vol.~60, pp.~1--14, 2021.

\bibitem{wang2022spcnet}
L.~Wang, L.~Wang, H.~Wang, X.~Wang, and L.~Bruzzone, ``Spcnet: A subpixel
  convolution-based change detection network for hyperspectral images with
  different spatial resolutions,'' {\em IEEE Transactions on Geoscience and
  Remote Sensing}, vol.~60, pp.~1--14, 2022.

\bibitem{tian2022racdnet}
J.~Tian, D.~Peng, H.~Guan, and H.~Ding, ``Racdnet: Resolution-and
  alignment-aware change detection network for optical remote sensing
  imagery,'' {\em Remote Sensing}, vol.~14, no.~18, p.~4527, 2022.

\bibitem{Liu2022g}
M.~Liu, Q.~Shi, A.~Marinoni, D.~He, X.~Liu, and L.~Zhang,
  ``{Super-Resolution-Based Change Detection Network With Stacked Attention
  Module for Images With Different Resolutions},'' {\em {IEEE} Transactions on
  Geoscience and Remote Sensing}, vol.~60, pp.~1--18, 2022.

\bibitem{liu2022learning}
M.~Liu, Q.~Shi, J.~Li, and Z.~Chai, ``Learning token-aligned representations
  with multimodel transformers for different-resolution change detection,''
  {\em IEEE Transactions on Geoscience and Remote Sensing}, vol.~60, pp.~1--13,
  2022.

\bibitem{chen2023remote}
H.~Chen, H.~Zhang, K.~Chen, C.~Zhou, S.~Chen, Z.~Zhou, and Z.~Shi, ``Continuous
  cross-resolution remote sensing image change detection,'' {\em IEEE
  Transactions on Geoscience and Remote Sensing}, 2023.

\bibitem{he2019spectral}
D.~He, Y.~Zhong, and L.~Zhang, ``Spectral--spatial--temporal map-based
  sub-pixel mapping for land-cover change detection,'' {\em IEEE Transactions
  on Geoscience and Remote Sensing}, vol.~58, no.~3, pp.~1696--1717, 2019.

\bibitem{heo2019comprehensive}
B.~Heo, J.~Kim, S.~Yun, H.~Park, N.~Kwak, and J.~Y. Choi, ``A comprehensive
  overhaul of feature distillation,'' in {\em Proceedings of the IEEE/CVF
  International Conference on Computer Vision}, pp.~1921--1930, 2019.

\bibitem{wang2021exploring}
W.~Wang, T.~Zhou, F.~Yu, J.~Dai, E.~Konukoglu, and L.~Van~Gool, ``Exploring
  cross-image pixel contrast for semantic segmentation,'' in {\em Proceedings
  of the IEEE/CVF International Conference on Computer Vision}, pp.~7303--7313,
  2021.

\bibitem{zhou2022rethinking}
T.~Zhou, W.~Wang, E.~Konukoglu, and L.~Van~Gool, ``Rethinking semantic
  segmentation: A prototype view,'' in {\em Proceedings of the IEEE/CVF
  Conference on Computer Vision and Pattern Recognition}, pp.~2582--2593, 2022.

\bibitem{yang2022cross}
C.~Yang, H.~Zhou, Z.~An, X.~Jiang, Y.~Xu, and Q.~Zhang, ``Cross-image
  relational knowledge distillation for semantic segmentation,'' in {\em
  Proceedings of the IEEE/CVF Conference on Computer Vision and Pattern
  Recognition}, pp.~12319--12328, 2022.

\bibitem{chen2020spatial}
H.~Chen and Z.~Shi, ``{A spatial-temporal attention-based method and a new
  dataset for remote sensing image change detection},'' {\em Remote Sensing},
  vol.~12, no.~10, p.~1662, 2020.

\bibitem{pang2023detecting}
C.~Pang, J.~Wu, J.~Ding, C.~Song, and G.-S. Xia, ``Detecting building changes
  with off-nadir aerial images,'' {\em Science China Information Sciences},
  vol.~66, no.~4, p.~140306, 2023.

\bibitem{CDDdataset}
M.~A. Lebedev, Y.~V. Vizilter, O.~V. Vygolov, V.~A. Knyaz, and A.~Y. Rubis,
  ``{CHANGE DETECTION IN REMOTE SENSING IMAGES USING CONDITIONAL ADVERSARIAL
  NETWORKS.},'' {\em International Archives of the Photogrammetry, Remote
  Sensing \& Spatial Information Sciences}, vol.~42, no.~2, 2018.

\bibitem{he2016deep}
K.~He, X.~Zhang, S.~Ren, and J.~Sun, ``{Deep residual learning for image
  recognition},'' in {\em CVPR}, pp.~770--778, 2016.

\bibitem{long2015fully}
J.~Long, E.~Shelhamer, and T.~Darrell, ``{Fully convolutional networks for
  semantic segmentation},'' in {\em ICCV}, pp.~3431--3440, 2015.

\bibitem{zhan2017change}
Y.~Zhan, K.~Fu, M.~Yan, X.~Sun, H.~Wang, and X.~Qiu, ``Change detection based
  on deep siamese convolutional network for optical aerial images,'' {\em IEEE
  Geoscience and Remote Sensing Letters}, vol.~14, no.~10, pp.~1845--1849,
  2017.

\bibitem{wang2020deep}
J.~Wang, K.~Sun, T.~Cheng, B.~Jiang, C.~Deng, Y.~Zhao, D.~Liu, Y.~Mu, M.~Tan,
  X.~Wang, {\em et~al.}, ``Deep high-resolution representation learning for
  visual recognition,'' {\em IEEE transactions on pattern analysis and machine
  intelligence}, vol.~43, no.~10, pp.~3349--3364, 2020.

\bibitem{Liang2022high}
Z.~Liang, B.~Zhu, and Y.~Zhu, ``{High resolution representation-based Siamese
  network for remote sensing image change detection},'' {\em IET Image Proc.},
  vol.~n/a, apr 2022.

\bibitem{chen2019spatioc}
C.~Chen and S.~Zhang, ``Spatio-temporal neural network with dilated
  retrospective convolution for short-time change detection,'' in {\em 2019 7th
  International Conference on Information, Communication and Networks (ICICN)},
  pp.~213--218, IEEE, 2019.

\bibitem{Zhang2020}
C.~Zhang, P.~Yue, D.~Tapete, L.~Jiang, B.~Shangguan, L.~Huang, and G.~Liu, ``{A
  deeply supervised image fusion network for change detection in high
  resolution bi-temporal remote sensing images},'' {\em ISPRS Journal of
  Photogrammetry and Remote Sensing}, vol.~166, pp.~183--200, aug 2020.

\bibitem{chen2020dasnet}
J.~Chen, Z.~Yuan, J.~Peng, L.~Chen, H.~Huang, J.~Zhu, Y.~Liu, and H.~Li,
  ``{DASNet: Dual Attentive Fully Convolutional Siamese Networks for Change
  Detection in High-Resolution Satellite Images},'' {\em IEEE Journal of
  Selected Topics in Applied Earth Observations and Remote Sensing}, vol.~14,
  pp.~1194--1206, 2021.

\bibitem{feng2022icif}
Y.~Feng, H.~Xu, J.~Jiang, H.~Liu, and J.~Zheng, ``Icif-net: Intra-scale
  cross-interaction and inter-scale feature fusion network for bitemporal
  remote sensing images change detection,'' {\em IEEE Transactions on
  Geoscience and Remote Sensing}, vol.~60, pp.~1--13, 2022.

\bibitem{fang2023changer}
S.~Fang, K.~Li, and Z.~Li, ``Changer: Feature interaction is what you need for
  change detection,'' {\em IEEE Transactions on Geoscience and Remote Sensing},
  2023.

\bibitem{shi2021deeply}
Q.~Shi, M.~Liu, S.~Li, X.~Liu, F.~Wang, and L.~Zhang, ``A deeply supervised
  attention metric-based network and an open aerial image dataset for remote
  sensing change detection,'' {\em IEEE transactions on geoscience and remote
  sensing}, vol.~60, pp.~1--16, 2021.

\bibitem{zheng2022changemask}
Z.~Zheng, Y.~Zhong, S.~Tian, A.~Ma, and L.~Zhang, ``{ChangeMask: Deep
  multi-task encoder-transformer-decoder architecture for semantic change
  detection},'' {\em ISPRS Journal of photogrammetry and remote sensing},
  vol.~183, pp.~228--239, jan 2022.

\bibitem{vaswani2017attention}
A.~Vaswani, N.~Shazeer, N.~Parmar, J.~Uszkoreit, L.~Jones, A.~N. Gomez,
  {\L}.~Kaiser, and I.~Polosukhin, ``{Attention is all you need},'' in {\em
  NIPS}, pp.~5998--6008, 2017.

\bibitem{swinsunet}
C.~Zhang, L.~Wang, S.~Cheng, and Y.~Li, ``{SwinSUNet: Pure transformer network
  for remote sensing image change detection},'' {\em IEEE Transactions on
  Geoscience and Remote Sensing}, vol.~60, pp.~1--13, 2022.

\bibitem{ledig2017photo}
C.~Ledig, L.~Theis, F.~Husz{\'a}r, J.~Caballero, A.~Cunningham, A.~Acosta,
  A.~Aitken, A.~Tejani, J.~Totz, Z.~Wang, {\em et~al.}, ``Photo-realistic
  single image super-resolution using a generative adversarial network,'' in
  {\em Proceedings of the IEEE conference on computer vision and pattern
  recognition}, pp.~4681--4690, 2017.

\bibitem{Shao2021}
R.~Shao, C.~Du, H.~Chen, and J.~Li, ``{{SUNet}: Change Detection for
  Heterogeneous Remote Sensing Images from Satellite and {UAV} Using a
  Dual-Channel Fully Convolution Network},'' {\em Remote Sensing}, vol.~13,
  p.~3750, jan 2021.

\bibitem{zheng2021unsupervised}
X.~Zheng, X.~Chen, X.~Lu, and B.~Sun, ``Unsupervised change detection by
  cross-resolution difference learning,'' {\em IEEE Transactions on Geoscience
  and Remote Sensing}, vol.~60, pp.~1--16, 2021.

\bibitem{he2020momentum}
K.~He, H.~Fan, Y.~Wu, S.~Xie, and R.~Girshick, ``Momentum contrast for
  unsupervised visual representation learning,'' in {\em Proceedings of the
  IEEE/CVF conference on computer vision and pattern recognition},
  pp.~9729--9738, 2020.

\bibitem{kamann2020benchmarking}
C.~Kamann and C.~Rother, ``Benchmarking the robustness of semantic segmentation
  models,'' in {\em Proceedings of the IEEE/CVF Conference on Computer Vision
  and Pattern Recognition}, pp.~8828--8838, 2020.

\bibitem{xie2021segformer}
E.~Xie, W.~Wang, Z.~Yu, A.~Anandkumar, J.~M. Alvarez, and P.~Luo, ``Segformer:
  Simple and efficient design for semantic segmentation with transformers,''
  {\em Advances in Neural Information Processing Systems}, vol.~34,
  pp.~12077--12090, 2021.

\bibitem{cui2022exploring}
Z.~Cui, Y.~Zhu, L.~Gu, G.-J. Qi, X.~Li, R.~Zhang, Z.~Zhang, and T.~Harada,
  ``Exploring resolution and degradation clues as self-supervised signal for
  low quality object detection,'' in {\em European Conference on Computer
  Vision}, pp.~473--491, Springer, 2022.

\bibitem{Fang2021}
S.~Fang, K.~Li, J.~Shao, and Z.~Li, ``{SNUNet-CD: A Densely Connected Siamese
  Network for Change Detection of VHR Images},'' {\em {IEEE} Geoscience and
  Remote Sensing Letters}, pp.~1--5, 2021.

\bibitem{Feng2023}
Y.~Feng, J.~Jiang, H.~Xu, and J.~Zheng, ``{Change Detection on Remote Sensing
  Images Using Dual-Branch Multilevel Intertemporal Network},'' {\em IEEE
  Transactions on Geoscience and Remote Sensing}, vol.~61, pp.~1--15, 2023.

\bibitem{zhang2016change}
P.~Zhang, M.~Gong, L.~Su, J.~Liu, and Z.~Li, ``Change detection based on deep
  feature representation and mapping transformation for
  multi-spatial-resolution remote sensing images,'' {\em ISPRS Journal of
  Photogrammetry and Remote Sensing}, vol.~116, pp.~24--41, 2016.

\end{thebibliography}
\end{document}